\documentclass[10pt,twocolumn,letterpaper]{article}

\usepackage[pagenumbers]{cvpr} % To force page numbers, e.g. for an arXiv version

\usepackage{graphicx}
\usepackage{amsmath}
\usepackage{amssymb}
\usepackage{booktabs}
\usepackage{bm}
\usepackage{physunits}

\usepackage{cuted}
\usepackage{capt-of}

\usepackage{multirow}
\usepackage{verbatim}
\usepackage{comment}
\usepackage{adjustbox}
\usepackage{wrapfig}

\usepackage{url}            % simple URL typesetting
\usepackage{amsfonts}       % blackboard math symbols
\usepackage{nicefrac}       % compact symbols for 1/2, etc.
\usepackage{microtype}      % microtypography

\usepackage{array}
\usepackage{xspace}
\usepackage{pifont}
\newcommand{\cmark}{\ding{51}}%
\newcommand{\xmark}{\ding{55}}%

\usepackage[dvipsnames]{xcolor}

\usepackage[accsupp]{axessibility}

\makeatletter
\renewcommand{\maketag@@@}[1]{\hbox{\m@th\normalsize\normalfont#1}}%
\makeatother

\newcommand\blfootnote[1]{%
  \begingroup
  \renewcommand\thefootnote{}\footnote{#1}%
  \addtocounter{footnote}{-1}%
  \endgroup
}

\definecolor{cvprblue}{rgb}{0.21,0.49,0.74}
\usepackage[pagebackref,breaklinks,colorlinks,citecolor=cvprblue]{hyperref}

\def\paperID{1311} % *** Enter the Paper ID here
\def\confName{CVPR}
\def\confYear{2024}

\title{I'M HOI: Inertia-aware Monocular Capture of 3D Human-Object Interactions}

\author{
    Chengfeng Zhao\textsuperscript{1}\qquad
    Juze Zhang\textsuperscript{1,2,3}\qquad
    Jiashen Du\textsuperscript{1}\qquad
    Ziwei Shan\textsuperscript{1}\qquad
    Junye Wang\textsuperscript{1}\\
    Jingyi Yu\textsuperscript{1}\qquad
    Jingya Wang\textsuperscript{1}\qquad
    Lan Xu\textsuperscript{1,$\dagger$}\\
    \textsuperscript{1}ShanghaiTech University \textsuperscript{2}Shanghai Advanced Research Institute, Chinese Academy of Sciences\\
    \textsuperscript{3}University of Chinese Academy of Sciences\\
    {\tt\footnotesize \{zhaochf2022,zhangjz,dujsh2022,shanzw2022,wangjy22022,yujingyi,wangjingya,xulan1\}@shanghaitech.edu.cn}
}

\begin{document}
\maketitle
\begin{strip}\centering
    \vspace{-45px}
    \captionsetup{type=figure}
    \includegraphics[width=\textwidth]{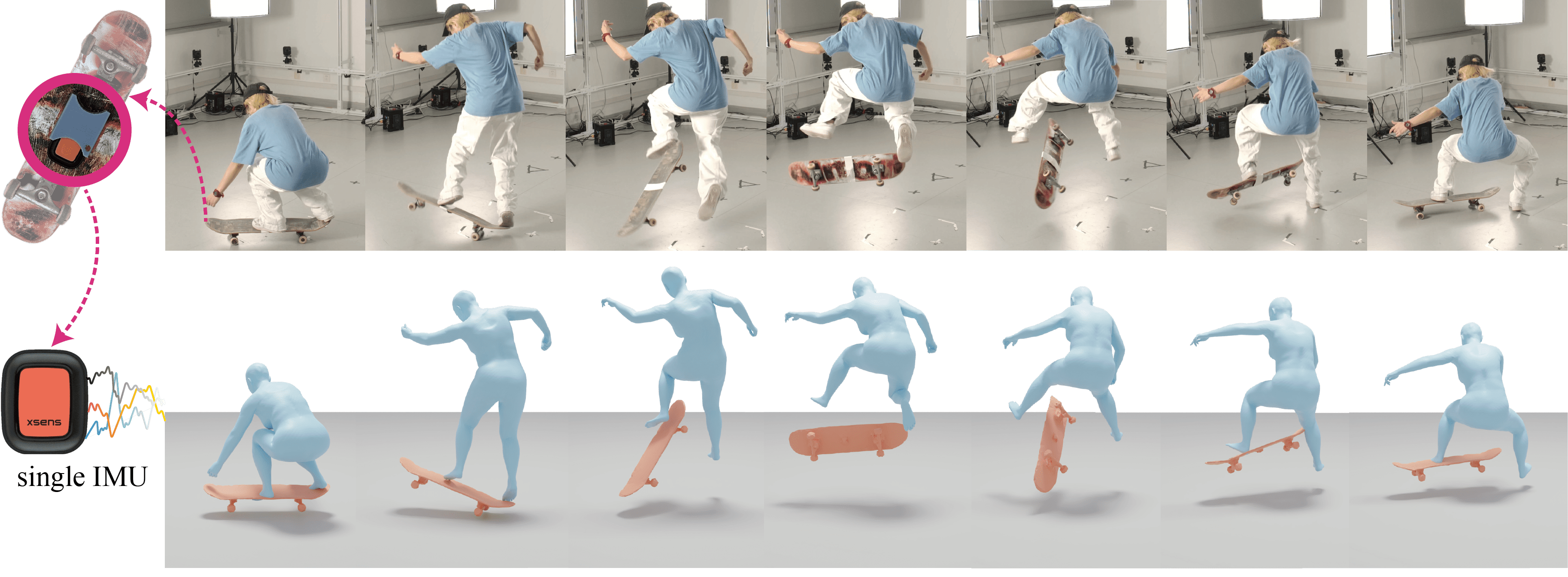}
    \vspace{-25px}
    \caption{Taking a monocular RGB video and a single inertial measurement unit (IMU) sensor recording, our approach, I'm-HOI, efficiently and robustly captures challenging and dynamic human-object interactions (HOI), such as skateboarding.}
    \label{fig:teaser}
    \vspace{-10px}
\end{strip}

\blfootnote{$^{\dagger}$Corresponding author}

% abstract
\begin{abstract}
% \vspace{-mm}
% 1. what's the task of our work? what's the significance(importance) of our work?
% 2. what's the current research status? what's the challenge?
% 3. to solve these problems, we propose ...
% 4. dataset
% 5. performance
We are living in a world surrounded by diverse and ``smart'' devices with rich modalities of sensing ability. Conveniently capturing the interactions between us humans and these objects remains far-reaching. In this paper, we present I'm-HOI, a monocular scheme to faithfully capture the 3D motions of both the human and object in a novel setting: using a minimal amount of RGB camera and object-mounted Inertial Measurement Unit (IMU). 
It combines general motion inference and category-aware refinement. For the former, we introduce a holistic human-object tracking method to fuse the IMU signals and the RGB stream and progressively recover the human motions and subsequently the companion object motions. For the latter, we tailor a category-aware motion diffusion model, which is conditioned on both the raw IMU observations and the results from the previous stage under over-parameterization representation. It significantly refines the initial results and generates vivid body, hand, and object motions. Moreover, we contribute a large dataset with ground truth human and object motions, dense RGB inputs, and rich object-mounted IMU measurements.
Extensive experiments demonstrate the effectiveness of I'm-HOI under a hybrid capture setting. Our dataset and code will be released to the community at the \href{https://afterjourney00.github.io/IM-HOI.github.io/}{project page}.
\end{abstract}    
\section{Introduction}

\label{sec:intro}

% 1. Convient capture human-object interactions is importtant
Capturing human-object interactions (HOI) is essential to understanding how we humans connect with the surrounding world, with numerous applications in robotics, gaming, or VR/AR. Yet, an accurate and convenient solution remains challenging in the vision community.

% 2. remained issues
% 2.1 multi-view: too heavy--> monocular, more attractive, yet cannot hand: occlusion and fast motions (skateboard) 
% 2.2 why multi-modality natual choice: novel and cute setting, most object are manual-made, easily intergrate various sensors 
% 2.3 various rgb+IMU, still not extend into HOI. Fundamentally, still lack dataset   
For high-fidelity capture of human-object interactions, early high-end solutions~\cite{bradley2008markerless,collet2015high,TotalCapture} require dense cameras, while recent approaches~\cite{motion2fusion,sun2021HOI-FVV,bhatnagar22behave,jiang2022neuralhofusion} require less RGB or RGBD video inputs (from 3 to 8 views). Yet, such a multi-view setting is still undesirable for consumer-level daily usage. Instead, the monocular method with more handiest captured devices is more attractive. Specifically, most recent methods~\cite{wang2022reconstruction,xie2022chore,zhang2020phosa,zhang2020object,xie2023vistracker} track the rigid and skeletal motions of objects and humans using a pre-scanned template or parametric model~\cite{SMPL2015} from a single RGB video input. Yet, inherently due to the RGB-setting, they remain vulnerable to depth ambiguity and the occlusion between human and object, especially for handling challenging fast motions like skateboarding. In contrast, the Inertial Measurement Units (IMUs) serve as a rescue to provide motion information that is robust to occlusion. Actually, IMU-based motion capture is widely adopted in both industry~\cite{XSENS} and academia~\cite{pons2011outdoor,huang2018DIP,TransPoseSIGGRAPH2021,ponton2023sparseposer}. Recent methods~\cite{liang2023hybridcap,pan2023fusing} further combine monocular RGB video and sparse IMUs, enabling lightweight and robust human motion capture. However, these schemes mostly focus on human-only scenarios and ignore the interacted objects. Moreover, compared to the sometimes tedious requirement of body-worn IMUs, it's more natural and convenient to attach the IMU sensor to the captured object, since IMUs have been widely integrated into daily objects like phones and smartwatches. Researchers surprisingly pay less attention to capturing human-object interactions with a minimal amount of RGB camera and IMU. The lack of motion data under rich interactions and modalities also constitute barriers to exploring such directions.

% 3. our key idea
% Explore a new setting: intertial-aid HOI, 
% a general capturing stage, and an catagery-aware motion stage
In this paper, we propose \textit{I'm-HOI} -- an inertia-aware and monocular approach for robustly tracking both the 3D human and object under challenging interactions (see Fig.~\ref{fig:teaser}). In stark contrast to prior arts, I'm-HOI adopts a lightweight and hybrid setting: a minimal amount of RGB camera and object-mounted IMU. Given the expected technological trend of mobile sensing as more and more RGB cameras and IMUs will be integrated into our surrounding devices, we believe that our approach will serve as a viable alternative to traditional human-object motion capture.

% 4. our technical pipeline
% 4.1 for the captureing stage, learning-optimization paradim
% 4.2 motion refineing, 
% 4.3 dataset: to train and evaluate our I'm-HOI, multi-modality dataset of human-object interactions
In I'm-HOI, our key idea is to adopt a two-stage paradigm to make full use of both the object-mounted IMU signals and the RGB stream, which consists of general motion inference and category-aware motion refinement.
For the former stage, we introduce holistic human-object tracking in an end-to-end manner. Specifically, we generate human motions via a multi-scale CNN-based network for 3D keypoints, followed by an Inverse Kinematics (IK) optimization layer. To reason the companion object motions, we progressively fuse the human features with IMU measurements via object-orientated mesh alignment feedback. We also adopt a robust optimization to refine the tracked object pose and improve the overlay performance, especially when the object is invisible in the RGB input. 
For the second refinement stage, we propose to tailor the conditional motion diffusion models~\cite{ho2020denoising,li2023ego} for utilizing category-level interaction priors. During training the diffusion model corresponding to a certain object, we treat the tracked motions and the raw IMU measurements from the previous stage as the condition information. We also adopt a novel over-parameterization representation with extra regularization designs to jointly consider the body, object, and especially hand regions during the denoising process. Thus, our refinement stage not only projects the initial human-object motions onto the category-specific motion manifold but also infills possible hand motions for vividly capturing human-object interactions.
To train and evaluate our I'm-HOI, we contribute a large multi-modal dataset of human-object interactions, covering 295 interaction sequences with 10 diverse objects in total 892k frames of recording. We also provide ground truth body, hand, and 3D object meshes, with dense RGB inputs and rich object-mounted IMU measurements.
% 5. our contribution
% new setting, 
% general capturing and catagory-awared diffusion for motion refinement, SOTA performance  
% new dataset, both our model and data will be released
To summarize, our main contributions include: 
\begin{itemize} 
    \setlength\itemsep{0em}
    \item We propose a multi-modal method to jointly capture human and object motions from a minimal amount of RGB camera and object-mounted IMU sensor.

    \item We adopt an efficient holistic human-object tracking method to progressively fuse the motion features, companion with a conditional diffusion model to refine and generate vivid interaction motions.
    
    \item We contribute a large dataset for human-object interactions, with rich RGB/IMU modalities and ground-truth annotations. Our data and model will be disseminated to the community.
	
\end{itemize}

\begin{figure*}[t!]
    \centering
    % \vspace{-5mm}
    \includegraphics[width=\textwidth]{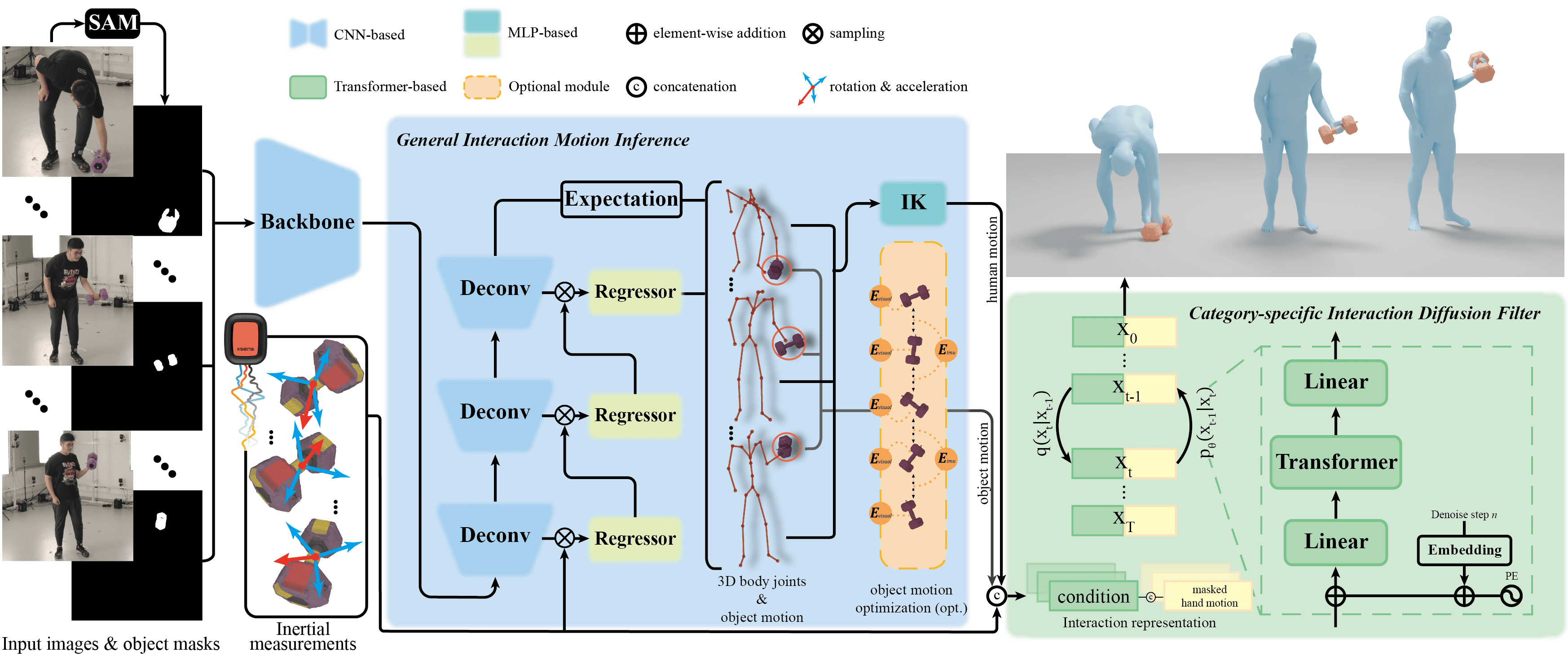}
    % \vspace{-6mm}
    \caption{The pipeline of I'm-HOI. Assuming video and inertial measurements input, our approach consists of a general interaction motion inference module (Sec.~\ref{sec:general}) and a category-specific interaction diffusion filter (Sec.~\ref{sec:specific}) to capture challenging interaction motions.}
    \label{fig:pipeline}
    % \vspace{-2mm}
\end{figure*}

\section{Related Work}

\label{sec:rw}

\paragraph{Monocular Human-centric Capture.} Since the release of the parametric body model SMPL~\cite{SMPL2015,MANO:SIGGRAPHASIA:2017,SMPLX2019}, there has been tremendous progress~\cite{keepitSMPL,Mehta2017,MonoPerfCap,HMR18,NBF:3DV:2018,pavlakos2018humanshape,tan2018indirect,Kolotouros_2019_CVPR,SPIN_ICCV2019,VIBE_CVPR2020,PHMR_ICCV2021,HUMOR_ICCV2021,PARE_ICCV2021,pymaf2021,li2021hybrik,zhang2023ikol,pymafx2023} in human motion capture from single RGB images and videos. However, reconstructing contextual human-object and human-scene interactions (HOI/HSI) from monocular input is far more challenging. The pioneer work PHOSA~\cite{zhang2020phosa} proposes a purely optimization-based framework to estimate static human-object spatial arrangements relying on handcrafted contact heuristics. This approach is unscalable and error-prone to depth ambiguities. Benefited from emerging 3D interaction motion datasets~\cite{pigraphs2016,PROX:2019,PSI:2019,caoHMP2020,hassan_samp_2021,bhatnagar22behave,huang2022intercap,huang2022rich,wang2022humanise,fan2023arctic,jiang2023fullbody}, learning-and-optimization work~\cite{xie2022chore,ijcai2023p100} has shown promising results by modeling human-object relative distance field in data-driven manner, followed by joint post-optimization. The state-of-the-art video-based method, VisTracker~\cite{xie2023vistracker}, further incorporates motion infilling techniques~\cite{yuan2022glamr} to enable space-time coherent tracking. However, these approaches still suffer from unacceptable runtime costs and unsatisfying accuracy under complex interaction scenarios.

\vspace{-4mm}
\paragraph{Inertial and Multi-modal Motion Capture.} Complementary to vision-based methods, human motion capture using inertial measurement units (IMUs) has also been extensively studied. Previous commercial solutions~\cite{XSENS,noitom} can capture accurate and detailed motion with dense sensors. Since the exploration of SIP~\cite{von2017SIP}, data-driven methods under sparse sensors configurations~\cite{huang2018DIP,TransPoseSIGGRAPH2021,PIPCVPR2022,TIP22,vanwouwe2023diffusion} have been developed to achieve real-time performance, deriving consumer-level products~\cite{mocopi}. To address the limitations of single-modal systems, multi-modal approaches~\cite{challencap} fuse inertial signals with RGB~\cite{pons2010multisensor,pons2011outdoor,von2016human,malleson2017real,trumble2017total,vonMarcard2018,gilbert2019fusing,henschel2020accurate,zhang2020fusing,malleson2020real,kaichi2020resolving,liang2023hybridcap,pan2023fusing}, RGBD~\cite{helten2013real,Zheng2018HybridFusion}, ego-view~\cite{EgoLocate2023}, and LiDAR~\cite{ren2023lip} references, achieving balanced local pose estimation and global localization. In this work, we extend the multi-sensor fusion strategy to 3D human-object interactions capture, which is beneficial for both accuracy and efficiency.

\vspace{-4mm}
\paragraph{Object-specific Interaction Prior.} Human motion priors have been demonstrated crucial for realism in the capture and synthesis by multiple modeling methodologies, including predefined kinematic structure~\cite{akhter2015pose,zhou2016deep}, GMM~\cite{SMPLX2019}, GAN~\cite{goodfellow2014generative,HMR18,BarsoumCVPRW2018}, VAE~\cite{kingma2013auto,HUMOR_ICCV2021,LEMO:ICCV:2021,petrovich21actor}, MLP~\cite{tiwari22posendf} and more cutting-edge, diffusion models~\cite{ho2020denoising,song2020denoising,tevet2023human,shafir2023human,yuan2023physdiff,karunratanakul2023gmd}. Additionally, context-aware human motion synthesis~\cite{Zhang:ICCV:2021,Zhao:ICCV:2023,mir23origin} and scene placement generation~\cite{yi2022mover,yi2023mime} successfully extract contextual prior knowledge from data. More recent work~\cite{petrov2023popup,xu2023interdiff,li2023object} modeled dynamic interaction patterns but ignored object category-level distribution differences. Other methods~\cite{zhang2022couch,kulkarni2023nifty} focus on specific interactions with chairs, which are static and lack diversity. This work aims to learn object category-specific interaction prior to model dynamic interaction distributions between human and diverse objects.
\section{Method} \label{sec:method}
We present a new paradigm for 3D human-object interactions capture in a lightweight and hybrid setting: utilizing a minimal amount of RGB camera and object-mounted Inertial Measurement Unit (IMU). 
As illustrated in Figure~\ref{fig:pipeline}, we propose a general interaction motion inference module (Sec.~\ref{sec:general}) to jointly recover human-object spatial arrangements in an end-to-end fashion. An category-specific interaction diffusion filter (Sec.~\ref{sec:specific}) is tailored to refine capture results from the former with the learned object category-level prior.

\subsection{General Interaction Motion Inference} \label{sec:general}
Current vision-based methods~\cite{xie2022chore,xie2023vistracker} typically adhere to fitting-learning-optimization framework, which we have observed to be susceptible to substantial or prolonged human-object occlusions, inefficient in inference time and limited in generalization capabilities, as discussed in Sec.~\ref{sec:comparison}. In contrast, we treat the object as an additional body joint and propose to estimate human-object spatial arrangements holistically and end-to-end. An optional optimization procedure can be incorporated to enhance capture accuracy further.
% Following off-the-shelf monocular single person motion capture methods~\cite{li2021hybrik,zhang2023ikol}, we first develop a CNN-based model to estimate 3D human body keypoints, together with the object geometry center. Subsequently, given the estimated 3D body joints, we recover human motion by solving the inverse kinematics problem. Assisted with the object-mounted IMU sensor, initially posed object with inertial rotation measurement and the estimated geometry center is fed to a coarse-to-fine group of regressors, supervised by mesh alignment feedback~\cite{pymaf2021,pymafx2023}, which is aimed to improve capture accuracy of object motion under frequent human-object occlusions. Additionally, an optional visual-inertial optimization procedure is designed to further enforce object motion comply with visual and inertial evidences.

\vspace{-4mm}
\paragraph{Preprocessing.} Given a monocular image sequence $\bm{I}\in \mathbb{R}^{T\times h\times w\times 3}$, we first segment human and object mask $\bm{S}_h,\bm{S}_o\in \mathbb{R}^{T\times h\times w\times 1}$ separately using SAM~\cite{kirillov2023segany}. Following that, a pre-trained ResNet-34~\cite{resnet_He2015} image encoder is adopted to extract image feature from stacked RGB image and object mask. After that, We take the raw inertial rotation $\bm{Q}\in \mathbb{R}^{T\times 6}$, acceleration $\bm{A}\in \mathbb{R}^{T\times 3}$ and normalized object template $\mathcal{O}$, combined with $\bm{I},\bm{S}_o$ as our network input. Our approach outputs human shape $\bm{\beta}\in\mathbb{R}^{T\times 10}$ and pose $\bm{\theta}\in\mathbb{R}^{T\times 3N_{J_b}}$, object rotation $\bm{R}_o\in\mathbb{R}^{T\times 6}$ and translation $\bm{T}_o\in\mathbb{R}^{T\times3}$. Here, $T=64$ is the sequence length, $h\times w$ is the resolution of images and masks, $N_{J_b}=22$ is the number of body joints. We adopt standard SMPL model~\cite{SMPL2015} for human motion representation. 

\vspace{-4mm}
\paragraph{End-to-end Holistic Human-Object Tracking.} We first introduce a multi-scale CNN-based network to jointly detect 3D human body joints $\bm{J}\in \mathbb{R}^{T\times 3N_{J_b}}$ and the object center. Leveraging the extracted image feature, we feed it into a series of deconvolution layers followed by a final convolution layer to reconstruct 3D keypoint heatmaps. The 3D keypoint positions are then determined by the expectation of each heatmap~\cite{sun2018integral}, with all keypoints canonicalized to root-relative representation except for the root joint. Commonly used combination of 3D keypoints and 2D reprojection loss $\mathcal{L}_{\text{kp3d}} + \lambda_{\text{j2d}}\mathcal{L}_{\text{j2d}}$ is utilized to train the CNN network, and simultaneously fine-tune the image extractor. Subsequent to 3D keypoint estimation, we employ and fine-tune an off-the-shelf pre-trained inverse kinematics layer~\cite{zhang2023ikol} by $\mathcal{L}_{\text{twist}}$ to recover human pose $\bm{\theta}$ and shape $\bm{\beta}$ based on $\hat{\bm{J}}$. Please refer to~\cite{xie2022chore,xie2023vistracker,li2021hybrik,zhang2023ikol} for detailed loss functions. 
 % We denote the overall 3D keypoints as $\bm{K}=[\bm{J},\bm{T}_o]\in \mathbb{R}^{T\times 3(N_{J_b}+1)}$.
% Taking the object as an extra body joint, we extend common 3D keypoints detection in human pose estimation to human-object interactions capture.
% The training loss is defined as:
% \begin{equation}
%     \mathcal{L}_{\text{kp}} = \lambda_{kp3d}\mathcal{L}_{\text{kp3d}} + \lambda_{\text{j2d}}\mathcal{L}_{\text{j2d}}.
%     \label{eq:kp_loss}
% \end{equation}
% Specifically, 3D supervision is directly performed as:
% \begin{equation}
%     \mathcal{L}_{\text{kp3d}} = \frac{1}{T}\sum_{t=0}^{T-1}||\hat{\bm{K}}_t - \bm{K}_t||_2^2,
%     \label{eq:kp3d_loss}
% \end{equation}
% and reprojected body joints loss:
% \begin{equation}
%     \mathcal{L}_{\text{j2d}} = \frac{1}{T}\sum_{t=0}^{T-1}\sum_{j=0}^{N_{J_b}-1}c^{(j)}||\mathcal{P}(\hat{\bm{J}}^{(j)}_t) - \bm{J}_{2D,t}^{(j)}||_2^2
%     \label{eq:kp2d_loss}
% \end{equation}
% is added for better human detections. In Equation~\ref{eq:kp2d_loss}, $\mathcal{P}(\cdot)$ is the camera projection function, $\bm{J}_{2D}$ is human body 2D annotation obtained from~\cite{OpenPose} and $c^{(j)}$ represents the confidence of the $j$-th joint.

For object tracking, we get an initially-posed object through $\mathcal{C}(\bm{Q})\mathcal{O}+\hat{\bm{T}_o}$ with estimated object translation and raw rotation data from the object-mounted IMU sensor, where $\mathcal{C}(\cdot)$ is the mapping from 6D rotation representation~\cite{zhou2019continuity} to rotation matrix. In order to eliminate systematic biases in $\bm{Q}$ and correct inaccurate $\hat{\bm{T}_o}$ under occlusions, we attach one MLP-based regressor to each intermediate image feature grid~\cite{pymaf2021,pymafx2023}, which forms a feedback loop to estimate corrective increment of object motion progressively. In the $i$-th loop, we first uniformly sample $N_S=400$ vertices on the posed object mesh $\mathcal{C}(\hat{\bm{R}}_o^{(i)})\mathcal{O} + \hat{\bm{T}}_o^{(i)}$. Then, we project sampled vertices onto the $i$-th feature map to obtain object mesh-aligned feature, which is subsequently fed into the $i$-th regressor to predict $\Delta\hat{\bm{R}}_o^{(i)}$ and $\Delta\hat{\bm{T}}_o^{(i)}$. Particularly, $\hat{\bm{R}}_o^{(0)} = \mathcal{C}(\bm{Q})$. The training loss of the feedback network group is defined as $\mathcal{L}_{\text{maf}} = \lambda_{\text{occ-sil}}\mathcal{L}_{\text{occ-sil}} + \lambda_{\text{area}}\mathcal{L}_{\text{area}}$, where $\mathcal{L}_{\text{occ-sil}}$ is the occlusion-aware silhouette loss proposed in~\cite{zhang2020phosa}. We find that better results can be achieved with an augmented silhouette area loss:
\begin{small}
\begin{equation}
    \mathcal{L}_{\text{area}} = \frac{1}{T}\sum_{t=0}^{T-1}\sum_{i=0}^{N_F-1}||\sum \mathcal{D}(\hat{\bm{R}}_{o,t}^{(i)}\mathcal{O}+\hat{\bm{T}}_{o,t}^{(i)}) - \sum\bm{S}_{o,t}||_2^2,
    \label{eq:area_loss}
\end{equation}
\end{small}where $\mathcal{D}(\cdot)$ refers to differentiable rendering function~\cite{ravi2020pytorch3d} and $N_F=3$ is the number of feedback iterations. Here, we re-define $\hat{\bm{R}}_o = \hat{\bm{R}}_o^{N_F-1}$ and $\hat{\bm{T}}_o = \hat{\bm{T}}_o^{N_F-1}$.

Overall, our training loss of the end-to-end inference module is:
\begin{equation}
    \mathcal{L} = \mathcal{L}_{\text{kp3d}} + \lambda_{\text{j2d}}\mathcal{L}_{\text{j2d}} + \mathcal{L}_{\text{twist}} + \mathcal{L}_{\text{maf}}.
\end{equation}
It's noteworthy that human mask $\bm{S}_{h}$ is only required in $\mathcal{L}_{\text{occ-sil}}$ during training but not when testing.
%correct initialized object motion iteratively.
%Inspired by~\cite{pymaf2021,pymafx2023},  we extend mesh alignment feedback from human pose estimation to object tracking in human-object interactions capture. To be specific, 
% The  resulting in $\hat{\bm{R}}_o^{(i+1)}=\hat{\bm{R}}_o^{(i)} + \Delta\hat{\bm{R}}_o^{(i)}$ and $\hat{\bm{T}}_o^{(i+1)}=\hat{\bm{T}}_o^{(i)} + \Delta\hat{\bm{T}}_o^{(i)}$.
% \begin{equation}
%     \mathcal{L}_{\text{maf}} = \lambda_{\text{occ-sil}}\mathcal{L}_{\text{occ-sil}} + \lambda_{\text{area}}\mathcal{L}_{\text{area}},
%     \label{eq:maf_loss}
% \end{equation}

\vspace{-4mm}
\paragraph{Robust and Lightweight Optimization.} To improve object tracking precision, especially in invisible cases, we propose an \textit{\textbf{optional}} optimization module. In addition to visual cues, we further constraint object rotation and trajectory to inertial measurements. The energy function is formulated as:
\begin{equation}
    \mathcal{E}=\mathcal{E}_{\text{visual}} + w_{\text{imu}}\mathcal{E}_{\text{imu}}.
\end{equation}
Specifically, $\mathcal{E}_{\text{visual}}$ minimizes the discrepancy between the rendering result and the segmentation of object: $\mathcal{E}_{\text{visual}} = \frac{1}{T}\sum_{t=0}^{T-1}\sum||\mathcal{D}(\hat{\bm{R}}_{o,t}\mathcal{O}+\hat{\bm{T}}_{o,t}) - \bm{S}_{o,t}||_2^2$. At the same time, $\mathcal{E}_{\text{imu}}$ regularizes object motion temporally:
\begin{equation}
    \begin{aligned}
        \mathcal{E}_{\text{imu}} = \frac{1}{T-1}\sum_{t=1}^{T-1}&||(\hat{\bm{T}}_{o,t-1} + \hat{\bm{T}}_{o,t+1} - 2\hat{\bm{T}}_{o,t}) - 0.5\bm{A}_t^2||_2^2 \\
        + \frac{1}{T}\sum_{t=0}^{T-1}&||\hat{\bm{R}}_{o,t} - \bm{Q}_t||_2^2.
    \end{aligned}
    \label{eq:inertial_energy}
\end{equation}

\subsection{Category-specific Interaction Diffusion Filter}
\label{sec:specific}
% specific interactions motion lies on different manifold depending on various functionalities and properties of objects. Second, hand-object interactions are highly frequent and subtle, taking a huge portion of human-object interactions. However, monocular hand motion capture is quite difficult due to factors such as low resolution, small visible region and tricky finger crossing.
In the second refinement stage, a category-specific interaction motion filter is proposed to \textit{(i)} project capture results from the preceding stage onto the manifold; \textit{(ii)} infill hand motions conditioned on body-object interaction motions.

% Some related work focusing on human-chair interactions~\cite{zhang2022couch,kulkarni2023nifty} take chairs' geometry as condition to synthesize realistic human motion. However, these methods only consider static and single-category objects. To overcome the limitations, we adopt a conditional diffusion model~\cite{ho2020denoising,song2020denoising} to formulate dynamic and category-specific human-object interactions prior. In addition to simple objectives~\cite{ho2020denoising}, we introduce several regularization terms to guide the learned manifold to be temporally consistent and physically plausible.

\vspace{-4mm}
\paragraph{Interaction Representation.}
We propose a novel over-parameterization interaction representation containing human, object motion and raw inertial measurements. At timestamp $t$ and noise level $n$, $\bm{x}_t^n \in \mathbb{R}^{486}$ consists of body-hand joint positions $\bm{j}_{h,t}\in \mathbb{R}^{156}$ and rotations $\bm{\theta}_{h,t}\in \mathbb{R}^{312}$; object translation $\bm{j}_{o,t} \in \mathbb{R}^{3}$ and rotation $\bm{\theta}_{o,t}\in \mathbb{R}^{6}$; inertial rotation $\bm{q}_t\in \mathbb{R}^6$ and free acceleration signal $\bm{a}_t\in \mathbb{R}^3$. We use 6D representation~\cite{zhou2019continuity} for all the rotation data, and the 52 joints body-hand model SMPL-H~\cite{MANO:SIGGRAPHASIA:2017} is adopted. The target interaction motion is represented as $\bm{x}_0$.

\begin{figure*}[t!]
    \centering
    \includegraphics[width=\textwidth]{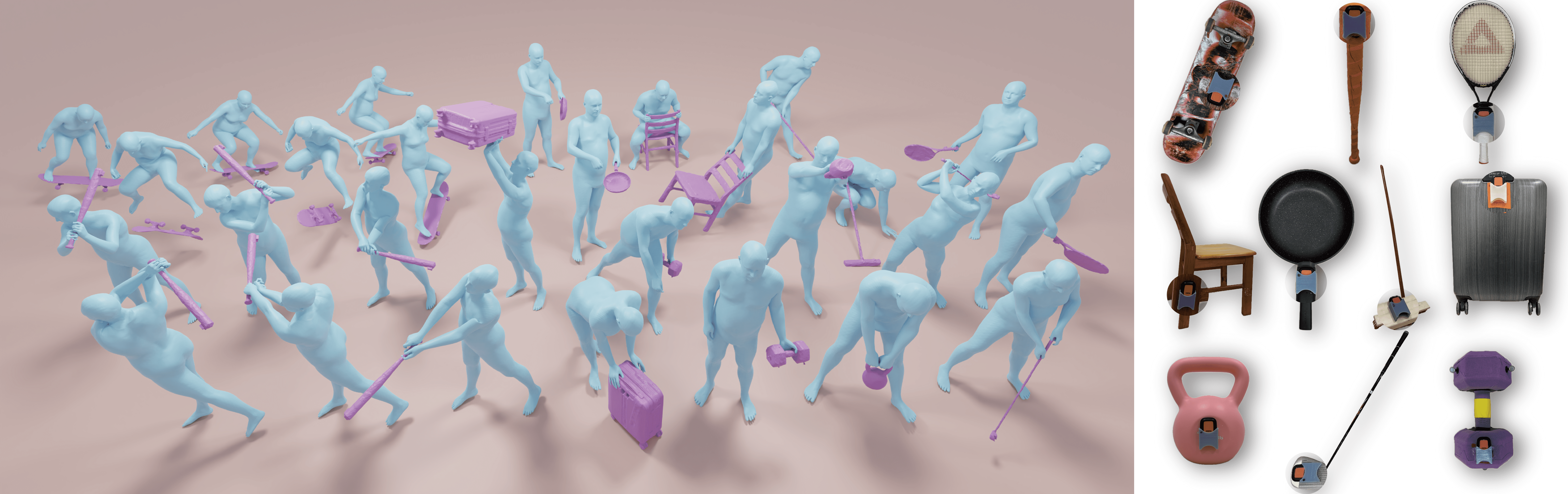}
    \caption{We exhibit selected highlights of IMHD$^2$ on the left side, and 10 well-scanned objects on the right side. In total, our dataset comprises 295 sequences and captures approximately 892k frames of data.}
    \label{fig:data}
\end{figure*}

\vspace{-4mm}
\paragraph{Conditional Diffusion Denoising Process.} Given $\bm{x}_0$, forward diffusion process adds Gaussian noise iteratively along an $N$-step Markov chain. For each noising step $n$, the noised interaction motion is drawn from conditional probability distribution determined by a pre-defined schedule $\{\alpha_n\}_{n=1}^{N}$:
\begin{equation}
    q(\bm{x}_{1:T}^n|\bm{x}_{1:T}^{n-1}) = \mathcal{N}(\sqrt{\alpha_n}\bm{x}_{1:T}^{n-1}, (1-\alpha_n)\mathcal{I}).
    \label{eq:forward}
\end{equation}
In reverse process, we formulate condition information as a tuple $\bm{c}=(\bm{j}_{h_b},\bm{j}_{o},\bm{\theta}_{h_b},\bm{\theta}_{o},\bm{q},\bm{a})\in \mathbb{R}^{216}$ and concatenate it with masked hand motion $\bm{m} = (\bm{j}_{h_h},\bm{\theta}_{h_h})\in \mathbb{R}^{270}$, where $\bm{j}_{h_b}\in \mathbb{R}^{66}$, $\bm{\theta}_{h_b}\in \mathbb{R}^{132}$ represents body-only joint positions and rotations. We follow~\cite{ramesh2022hierarchical} to predict $\bm{x}_0$ itself as $\hat{\bm{x}}_{\phi}(\bm{x}_{n},n,\bm{c})$, where $\phi$ represents the parameters of neural network. The training loss is $L1$-norm simple objective~\cite{ho2020denoising,li2023ego}:
\begin{equation}
    \mathcal{L}_{\text{simple}} = \mathbb{E}_{\bm{x}_0, n}||\hat{\bm{x}}_{\phi}(\bm{x}_{n}, n, \bm{c}) - \bm{x}_{0}||_{1}.
    \label{eq:diffusion_loss}
\end{equation}
Inspired by~\cite{HUMOR_ICCV2021}, components inside of the over-parameterization representation have mutual constraints. We accordingly introduce four regularization terms: 
\begin{equation}
    \mathcal{L}_{\text{reg}} = \lambda_{\text{off}}\mathcal{L}_{\text{off}} + \lambda_{\text{vel}}\mathcal{L}_{\text{vel}} + \lambda_{\text{consist}}\mathcal{L}_{\text{consist}} + \lambda_{\text{imu}}\mathcal{L}_{\text{imu}}.
    \label{{eq:regularization}}
\end{equation}
Specifically, $\mathcal{L}_{\text{off}}$ enforces the predicted object center to lie in a small region determined by the distance offsets relative to 52 body-hand joints as:
\begin{equation}
    \mathcal{L}_{\text{off}} = \frac{1}{T}\sum_{t=0}^{T-1}\sum_{i=0}^{N_J}||(\hat{\bm{j}}_{o,t}-\hat{\bm{j}}_{h,t}^{(i)}) - (\bm{j}_{o,t}-\bm{j}_{h,t}^{(i)})||_1.
    \label{eq:offset_loss}
\end{equation}
We then constraint the reproduced human body-hand joints consistent with the body model skinned from predicted joint rotations:
\vspace{-1.5mm}
\begin{equation}
    \mathcal{L}_{\text{consist}} = \frac{1}{T}\sum_{t=0}^{T-1}||\hat{\bm{j}}_{h,t} - \mathcal{J}(\mathcal{M}(\hat{\bm{\beta}}, \hat{\bm{\theta}}_{h,t}))||_1,
    \label{eq:consist_loss}
\end{equation}

\vspace{-1.5mm}
\noindent where $\mathcal{M}(\cdot)$ refers to forward function of SMPL-H model and $\mathcal{J}(\cdot)$ is the joint regressor. In order to temporally smooth human motion, the velocity term $\mathcal{L}_{\text{vel}}$ is formulated as:
\vspace{-1.5mm}
\begin{equation}
    \mathcal{L}_{\text{vel}} = \frac{1}{T-1}\sum_{t=1}^{T-1}||(\hat{\bm{j}}_{t} - \hat{\bm{j}}_{t-1}) - (\bm{j}_{t} - \bm{j}_{t-1})||_1,
    \label{eq:velocity_loss}
\end{equation}

\vspace{-1.5mm}
\noindent where $\bm{j}_{t} = [\bm{j}_{h,t}, \bm{j}_{o,t}]$. Finally, $\mathcal{L}_{\text{imu}}$ guides the generated object poses and the trajectory conform to IMU measurements, which improves robustness under invisible scenarios:
\vspace{-1.5mm}
\begin{equation}
    \mathcal{L}_{\text{imu}} = \mathcal{L}_{\text{rot}} + \frac{1}{T-1}\sum_{t=1}^{T-1}\mathcal{L}_{\text{acc},t}.
    \label{eq:imu_loss}
\end{equation}

\vspace{-1.5mm}
\noindent Wherein, object rotation is directly regularized by:
\vspace{-1.5mm}
\begin{equation}
    \mathcal{L}_{\text{rot}} = \frac{1}{T}\sum_{t=0}^{T-1}||\hat{\bm{\theta}}_{o,t} - \bm{q}_{t}||_1,
    \label{eq:rot_loss}
\end{equation}

\vspace{-1.5mm}
\noindent and the trajectory is constrained to acceleration through:
\vspace{-1.5mm}
\begin{equation}
    \mathcal{L}_{\text{acc},t} = ||(\hat{\bm{j}}_{o,t}-\hat{\bm{j}}_{o,t-1} + \frac{\bm{a}_t\tau^2}{2}) - (\bm{j}_{o,t+1} - \bm{j}_{o,t})||_1,
    \label{eq:acc_loss}
\end{equation}

\vspace{-1.5mm}
\noindent where $\tau$ is the time interval between two consecutive frames. It's noteworthy that, related work~\cite{xie2023vistracker} simulates such second-order constraints in a pseudo manner~\cite{zeng2022smoothnet} to eliminate mutation of first-order signals. In contrast, we incorporate acceleration explicitly~\cite{liang2023hybridcap,ren2023lip}.
\begin{figure*}[th!]
    \centering
    \includegraphics[width=\textwidth]{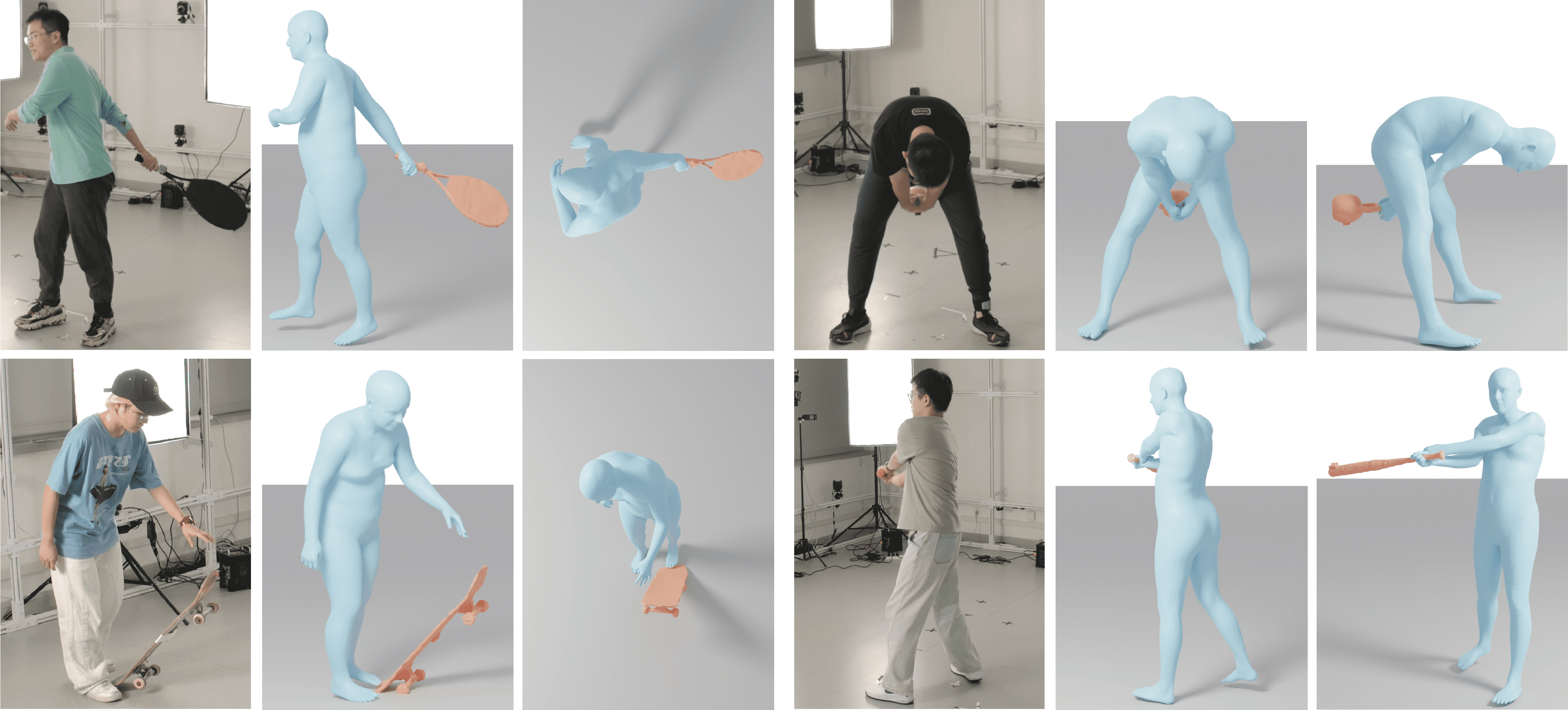}
    \caption{Qualitative 3D capturing results of I'm-HOI on IMHD$^2$ dataset. Each sample includes an RGB image input, captured motion from camera view, and side-view visualization.}
    \label{fig:result_gallery}
\end{figure*}

\begin{table*}[t!]
    \centering
    \small
    \setlength{\tabcolsep}{7pt}
    \begin{tabular}{lcccccccccccc}
        \toprule
         & \multicolumn{4}{c}{IMHD$^2$ (fast)} & \multicolumn{4}{c}{BEHAVE (slow to medium)~\cite{bhatnagar22behave}} & \multicolumn{4}{c}{InterCap (slow)~\cite{huang2022intercap}} \\
        \cmidrule(lr){2-5} \cmidrule(lr){6-9} \cmidrule(lr){10-13}
         & \multicolumn{2}{c}{CD (per-frame)} & \multicolumn{2}{c}{CD ($10s$)} & \multicolumn{2}{c}{CD (per-frame)} & \multicolumn{2}{c}{CD ($10s$)} & \multicolumn{2}{c}{CD (per-frame)} & \multicolumn{2}{c}{CD ($10s$)} \\
         \cmidrule(lr){2-3} \cmidrule(lr){4-5} \cmidrule(lr){6-7} \cmidrule(lr){8-9} \cmidrule(lr){10-11} \cmidrule(lr){12-13}
         Method & smpl & object & smpl & object & smpl & object & smpl & object & smpl & object & smpl & object\\
        \midrule
        PHOSA~\cite{zhang2020phosa} & 29.20 & 20.26 & 41.26 & 56.80 & 12.86 & 26.90 & 27.01 & 59.08 & 11.20 & 20.57 & 24.16 & 43.06 \\
        CHORE~\cite{xie2022chore} & 14.20 & 16.81 & 24.32 & 31.76 & 5.55 & 10.02 & 18.33 & 20.32 & 7.12 & 12.59 & 16.11 & 21.05 \\
        VisTracker~\cite{xie2023vistracker} & 19.96 & 23.28 & 17.02 & 18.10 & \textbf{5.25} & 8.04 & 7.81 & 8.49 & 6.76 & 10.32 & 9.35 & 11.38 \\
        Ours & \textbf{6.50} & \textbf{6.93} & \textbf{5.36} & \textbf{8.53} & 5.26 & \textbf{7.43} & \textbf{5.65} & \textbf{4.82} & \textbf{5.66} & \textbf{8.92} & \textbf{5.81} & \textbf{7.14} \\
        \bottomrule
    \end{tabular}
    \caption{Quantitative comparison was conducted with several baselines on both human and object tracking accuracy.}
    \label{tab:comparison}
\end{table*}

\section{Dataset}
\label{sec:dataset}
To train and evaluate our I'm-HOI, we collect an \textit{\textbf{I}nertial and \textbf{M}ulti-view \textbf{H}ighly \textbf{D}ynamic human-object interactions \textbf{D}ataset} (\textit{\textbf{IMHD$^2$}}), consisting of human, object motions, inertial measurements and object 3D scans.

\vspace{-4mm}
\paragraph{Capture Preparations.} A high-end multi-view camera system consisting of 32 Z CAMs~\cite{Zcam} was set up to capture 4K videos at 60 fps. Simultaneously, two Xsens DOT IMU sensors~\cite{XSENS} mounted on the object and the leg of performer were used to record object inertia and align timestamps at 60 $\Hz$ respectively. We invited 15 subjects (13 males, 2 females) to participate in 10 different interaction scenarios. Sequence-level textual guidance was provided for each capture split to ensure reasonable and meaningful interactions. Each split lasted from half a minute to one minute. We conducted visual-inertial system calibration once per ten minutes to eliminate disturbances caused by magnetic field changes.

\vspace{-4mm}
\paragraph{Data Processing.} Given multi-view videos, we reproduced human motions in SMPL-H format~\cite{MANO:SIGGRAPHASIA:2017} using an open-source toolbox~\cite{easymocap}. To accurately track object pose in a 3D scene, we manually annotated single key-frame segmentation in all views and broadcasted it to the entire sequence~\cite{yang2023track,kirillov2023segany,cheng2022xmem,liCvpr22vInpainting}. Subsequently, we optimized Euclidean transformations, which precisely define object motions, by fitting reprojected silhouette to multi-view masks. For object geometries, we utilized a public application~\cite{Polycam} to obtain 3D scan templates. In terms of inertial signals, we adopted primitive rotation data $\bm{R}_s$ in matrix form and transformed raw acceleration $\bm{a}_{raw}$ in sensor coordinate to free acceleration $\bm{a}_{free}$ in global coordinate through $\bm{a}_{free} = \bm{R}_s\bm{a}_{raw} - \bm{g}$, where $\bm{g} = [0, 0, 9.81]^{\mathrm{T}}$ is the gravitational acceleration.
\begin{figure*}[t!]
    \centering
    \includegraphics[width=0.9\textwidth]{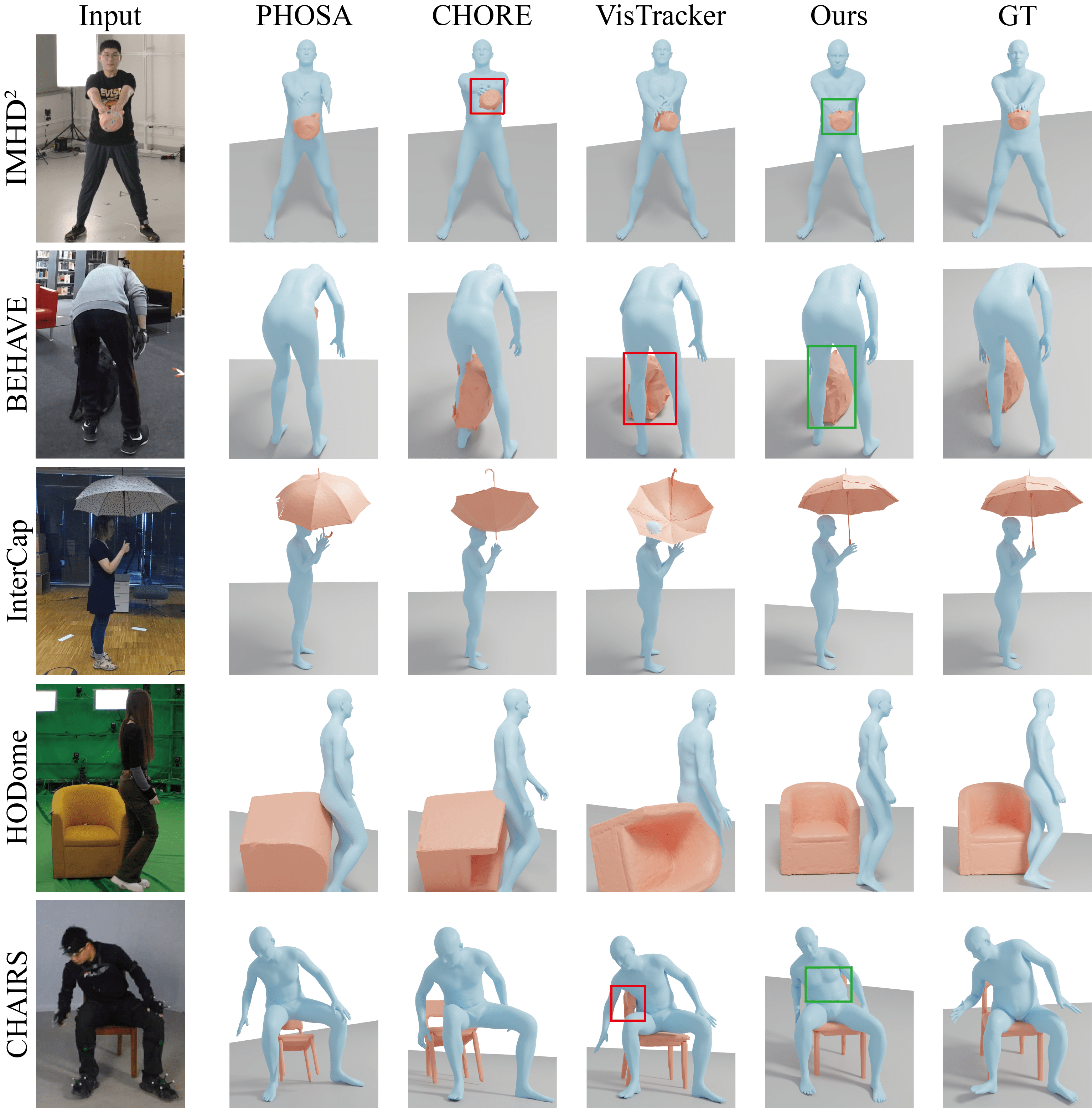}
    \caption{Qualitative comparison results. I'm-HOI outperforms the baselines and generalizes well to new datasets.}
    \label{fig:comparison}
    % \vspace{-5mm}
\end{figure*}

\section{Experiments}

\label{sec:exp}

In this section, we first introduce the datasets and metrics used for training and evaluation. We then provide a comprehensive comparison between our approach with baseline methods. We also perform extensive ablation studies to demonstrate the effectiveness of pivot components in our network design and the necessity of the IMU modality.

\subsection{Datasets and Evaluation Metrics}

We train I'm-HOI using BEHAVE~\cite{bhatnagar22behave}, InterCap~\cite{huang2022intercap} and IMHD$^2$, and evaluate it on five datasets which also include HODome~\cite{zhang2023neuraldome} and CHAIRS~\cite{jiang2023fullbody}. We adhere to the official train-test data partitioning of BEHAVE and InterCap, which is established by VisTracker~\cite{xie2023vistracker}. Given the relatively slow inference speeds of baselines~\cite{zhang2020phosa,xie2022chore,xie2023vistracker}, we curate partial yet representative data from IMHD$^2$, HODome, and CHAIRS to construct sub test sets for thorough evaluation.

\vspace{-4mm}
\paragraph{Evaluation Metrics.}
\begin{itemize}
    \item \textbf{Per-frame Chamfer Distance ($cm$)}~\cite{xie2022chore} computes the chamfer distance between predicted human and object mesh with the ground truth respectively after holistic Procrustes alignment for every single frame.
    \item \textbf{Sliding Window Chamfer Distance ($cm$)}~\cite{xie2023vistracker} computes the chamfer distance in the same way but performing holistic Procrustes alignment on the combined mesh of 10-second results with the ground truth.
\end{itemize}

\subsection{Comparison} \label{sec:comparison}
\paragraph{Results.} 
As shown in Table~\ref{tab:comparison}, I'm-HOI consistently outperforms the baselines on several datasets, especially on IMHD$^2$ which is characterized by fast interaction motions, with a large margin around 15$cm$. We visualize qualitative results in Figure~\ref{fig:comparison}, I'm-HOI captures better human-object spatial arrangements, including both relative pose and position. In addition, our approach shows better robustness than baselines under severe occlusions.

\vspace{-4mm}
\paragraph{Generalization.} To assess the generalization capabilities, we evaluate the performance of purely optimization-based method PHOSA~\cite{zhang2020phosa}, learning-and-optimization methods CHORE\cite{xie2022chore} and VisTracker\cite{xie2023vistracker}, as well as our proposed approach trained on BEHAVE\cite{bhatnagar22behave} and InterCap\cite{huang2022intercap}, across HODome\cite{zhang2023neuraldome} and CHAIRS\cite{jiang2023fullbody}. As shown in Table~\ref{tab:generalization}, I'm-HOI generalizes better than the baselines by a large margin and achieves more balanced performance between per-frame and sequential results. Furthermore, Figure~\ref{fig:comparison} demonstrates the adaptability to diverse scenarios of I'm-HOI.

\vspace{-4mm}
\paragraph{Runtime Cost.} We conduct a comparative analysis of the inference efficiency across different methods using a specific sequence from InterCap dataset~\cite{huang2022intercap}. Among the methods evaluated, the purely optimization-based framework PHOSA~\cite{zhang2020phosa} takes the longest inference time which is approximately 2 minutes per frame. CHORE~\cite{xie2022chore} speeds up to 1 minute per frame, while VisTracker~\cite{xie2023vistracker} further reduces the time cost to 20 seconds. Notably, I'm-HOI requires only about 0.5 seconds per frame for the complete pipeline. It is worth mentioning that omitting the optional optimization module could lead to additional enhancements in efficiency.

\subsection{Ablation Study}
Extensive ablation studies are conducted on IMHD$^2$ to evaluate our network architecture design and IMU modality.

\begin{figure}[t!]
    \centering
    \includegraphics[width=0.47\textwidth]{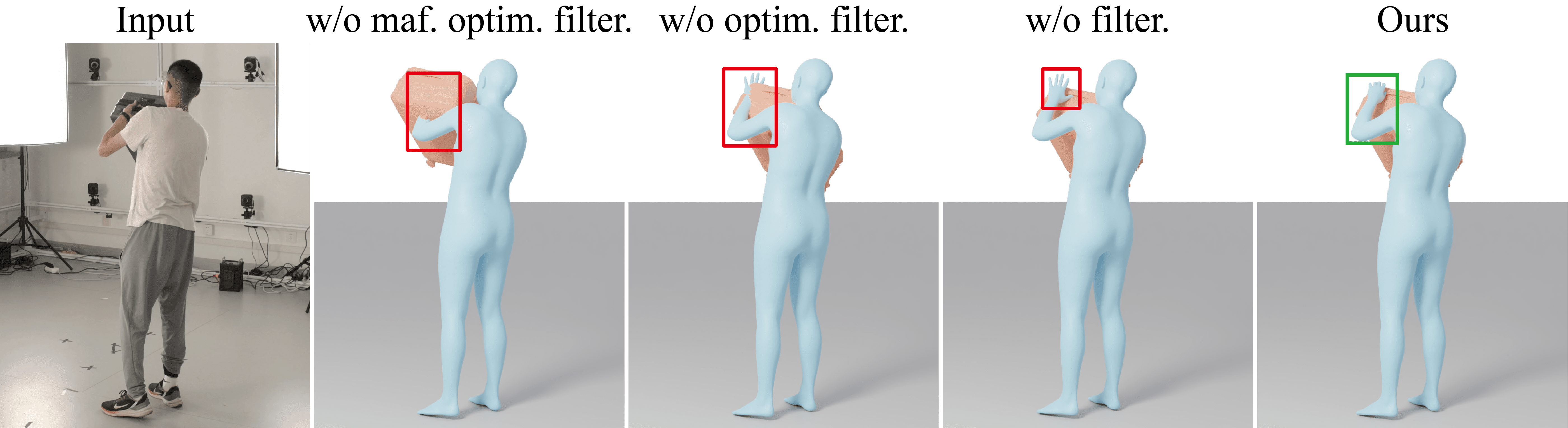}
    \vspace{-2mm}
    \caption{Qualitative evaluation of our network architecture. The figure illustrates the effectiveness of each key design.}
    \label{fig:as_na}
    \vspace{-4mm}
\end{figure}

\vspace{-4mm}
\paragraph{Network Architecture.} Table~\ref{tab:eval_na} shows the performance of models with and without the mesh alignment feedback (maf.), optimization module (optim.) and diffusion filter (filter.). It is demonstrated that maf. improves per-frame object tracking results and optim. brings better temporal consistency. In addition, filter. further corrects human-object spatial arrangements onto the learned interaction manifold. Compared to the naive implementation, the full pipeline of I'm-HOI performs 4 times better. Figure~\ref{fig:as_na} illustrates that inaccurate prediction is progressively corrected when maf. and optim. are applied. Also, the hand motion generated by filter. makes the capture result more vivid and realistic.

\begin{table}[t!]
    \centering
    \footnotesize
    \setlength{\tabcolsep}{4.9pt}
    \renewcommand{\arraystretch}{0.9}
    \begin{tabular}{llcccc}
        \toprule
         & & \multicolumn{2}{c}{CD (per-frame)} & \multicolumn{2}{c}{CD ($10s$)}\\
        \cmidrule(lr){3-4} \cmidrule(lr){5-6}
        Dataset & Method & smpl & object & smpl & object \\
        \midrule
        \multirow{4}{*}{HODome~\cite{zhang2023neuraldome}} & PHOSA~\cite{zhang2020phosa} & 34.41 & 29.70 & 60.15 & 54.98 \\
        & CHORE~\cite{xie2022chore} & 23.18 & 16.18 & 43.35 & 31.64 \\
        & VisTracker~\cite{xie2023vistracker} & 11.87 & 19.86 & 32.77 & 34.53 \\
        & Ours & \textbf{8.19} & \textbf{9.05} & \textbf{12.07} & \textbf{15.31} \\
        \midrule
        \multirow{4}{*}{CHAIRS~\cite{jiang2023fullbody}} & PHOSA~\cite{zhang2020phosa} & 35.26 & 28.35 & 43.17 & 37.67 \\
        & CHORE~\cite{xie2022chore} & 19.10 & 36.13 & 16.71 & 52.95 \\
        & VisTracker~\cite{xie2023vistracker} & 17.42 & 23.31 & 15.23 & 16.85 \\
        & Ours & \textbf{9.55} & \textbf{9.91} & \textbf{6.34} & \textbf{7.85} \\
        \bottomrule
  \end{tabular}
  \vspace{-2mm}
  \caption{Quantitative evaluations of generalization ability.}
  \label{tab:generalization}
  \vspace{-2mm}
\end{table}

\begin{table}[t!]
    \centering
    \small
    \setlength{\tabcolsep}{6.5pt}
    \renewcommand{\arraystretch}{0.9}
    \begin{tabular}{lllcccc}
        \toprule
        \multicolumn{3}{l}{Module} & \multicolumn{2}{c}{CD (per-frame)} & \multicolumn{2}{c}{CD ($10s$)}\\
        \cmidrule(r){1-3} \cmidrule(lr){4-5} \cmidrule(l){6-7}
        maf. & optim. & filter. & smpl & object & smpl & object \\
        \midrule
        \textcolor{red}{\xmark} & \textcolor{red}{\xmark} & \textcolor{red}{\xmark} & 8.42 & 15.12 & 13.88 & 27.85 \\
        \textcolor{green}{\cmark} & \textcolor{red}{\xmark} & \textcolor{red}{\xmark} & 8.02 & 9.73 & 10.31 & 19.87 \\
        \textcolor{red}{\xmark} & \textcolor{red}{\xmark} & \textcolor{green}{\cmark} & 37.35 & 35.14 & 48.90 & 65.03 \\
        \textcolor{green}{\cmark} & \textcolor{green}{\cmark} & \textcolor{red}{\xmark} & 7.16 & 7.42 & 7.25 & 10.75 \\
        \textcolor{green}{\cmark} & \textcolor{red}{\xmark} & \textcolor{green}{\cmark} & 7.52 & 8.80 & 8.56 & 12.41\\
        \textcolor{green}{\cmark} & \textcolor{green}{\cmark} & \textcolor{green}{\cmark} & \textbf{6.50} & \textbf{6.93} & \textbf{5.36} & \textbf{8.53} \\
        \bottomrule
  \end{tabular}
  \vspace{-2mm}
  \caption{Quantitative evaluation of network architecture design.}
  \label{tab:eval_na}
  \vspace{-2mm}
\end{table}

\begin{figure}[t!]
    \centering
    \includegraphics[width=0.47\textwidth]{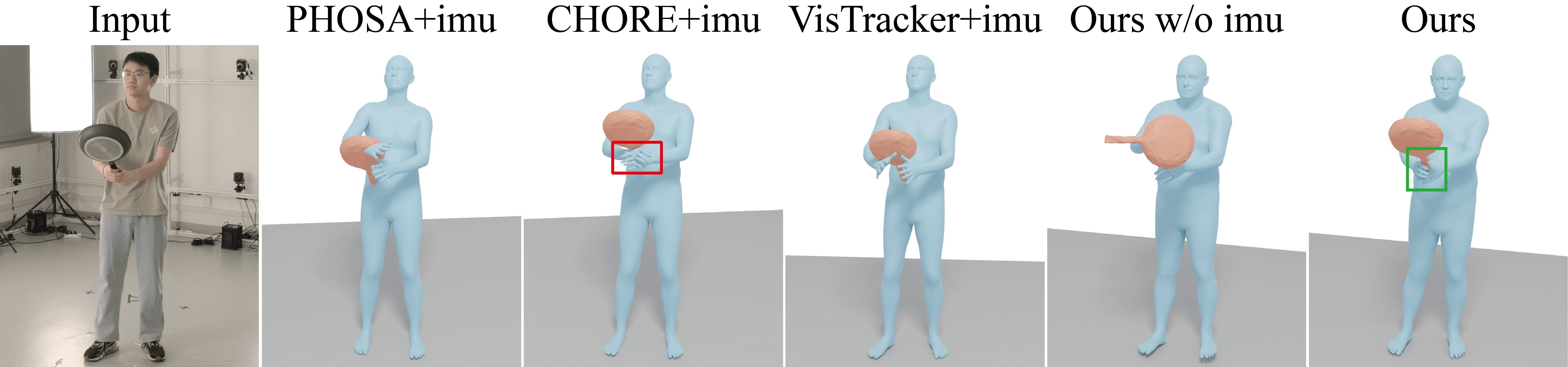}
    \vspace{-2mm}
    \caption{Qualitative evaluation of the IMU modality. This figure shows the importance of inertial measurements input.}
    \label{fig:as_im}
    \vspace{-4mm}
\end{figure}

\vspace{-4mm}
\paragraph{Input Modality.} We experiment on several different baselines with IMU modality input by adding an additional inertial optimization term described in Equation~\ref{eq:inertial_energy} to their pipelines. The qualitative results shown in Figure~\ref{fig:as_im} clearly demonstrate the improvements in object pose estimation of the baselines after introducing the IMU modality, compared to Figure~\ref{fig:comparison}. The quantitative performance reported in Table~\ref{tab:eval_im} shows a decent increase in the performance of baselines, compared to the statistics in Table~\ref{tab:comparison}. Additionally, we observe that our approach achieves better results when the IMU modality is involved, especially for object tracking. Furthermore, Table~\ref{tab:eval_im} shows that naively incorporating the IMU modality input into baselines is unable to maximize its benefits, which further verifies the effectiveness of our network design.

\subsection{Limitation}
The proposed I'm-HOI is the first trial to explore challenging 3D human-object interactions capture using a minimal amount of RGB camera and object-mounted IMU sensor. However, it still has limitations. Firstly, our method relies on pre-scanned object templates and manual manipulation of sensor-template coordinate alignment. Additionally, our method is restricted to rigid object tracking. Extending this method to articulated or even deformable objects in a template-free framework is promising.
\begin{table}[t!]
    \centering
    \small
    \setlength{\tabcolsep}{9pt}
    \renewcommand{\arraystretch}{0.9}
    \begin{tabular}{lcccc}
        \toprule
        & \multicolumn{2}{c}{CD (per-frame)} & \multicolumn{2}{c}{CD ($10s$)}\\
        \cmidrule(lr){2-3} \cmidrule(l){4-5}
        Modality & smpl & object & smpl & object \\
        \midrule
        PHOSA+imu & 28.41 & 18.60 & 39.19 & 38.44 \\
        CHORE+imu & 12.98 & 11.92 & 22.09 & 23.31 \\
        VisTracker+imu & 15.87  & 14.61 & 14.68 & 12.82 \\
        \midrule
        Ours w/o imu & 10.58 & 14.49 & 6.55 & 15.23 \\
        Ours & \textbf{6.50} & \textbf{6.93} & \textbf{5.36} & \textbf{8.53} \\
        \bottomrule
  \end{tabular}
  \vspace{-2mm}
  \caption{Quantitative evaluations on input modality configurations.}
  \label{tab:eval_im}
  \vspace{-2mm}
\end{table}

\section{Conclusion}
We have presented a novel and monocular scheme to faithfully capture the 3D motions of human-object interactions, using a minimal amount of RGB camera and object-mounted IMU. 
Our general motion inference stage progressively fuses the IMU signals and the RGB stream via holistic and end-to-end tracking, which efficiently recovers the human motions and subsequently the companion object motions via mesh alignment feedback.  Our category-aware motion diffusion further treats the previous results as conditions and jointly considers the body, object, and especially hand regions during the denoising process with an over-parameterization representation. It encodes category-aware motion priors, so as to significantly improve the tracking accuracy and generate vivid hand motions.
Our experimental results demonstrate the effectiveness of I'm-HOI for faithfully capturing human and object motions in a lightweight setting. As more and more sensors like RGB cameras or IMUs will be integrated into our surrounding world, we believe that our approach and dataset will serve as a critical step towards hybrid human-object motion capture,  with many potential applications in robotics, embodied AI, or VR/AR. 

\vspace{-4mm}
\paragraph{Acknowledgement.} This work was supported by National Key R\&D Program of China (2022YFF0902301), Shanghai Local college capacity building program (22010502800). We also acknowledge support from Shanghai Frontiers Science Center of Human-centered Artificial Intelligence (ShangHAI).

% \newpage
{
    \small
    \bibliographystyle{ieeenat_fullname}
    \bibliography{main}
}

% WARNING: do not forget to delete the supplementary pages from your submission 
\clearpage
\maketitlesupplementary
\appendix

In this supplementary, we commence by providing a comprehensive exposition on the implementation details of our method. Subsequently, we expound upon the calibration of the multi-modal system and the processing of IMU acceleration data during the data collection phase of IMHD$^2$. Finally, we present additional qualitative and quantitative results to further validate the efficacy of I'm-HOI, with a particular focus on the evaluation of each regularization term integrated within the category-specific interaction diffusion filter component.

\section{Implementation Details}

\subsection{General Interaction Motion Inference}

\paragraph{Network Architecture.} We employ a pre-trained ResNet-34 model~\cite{resnet_He2015} $f^{\text{enc}}:\mathbb{R}^{H\times W\times 4}\mapsto \mathbb{R}^{8\times 8\times 512}$ to extract image features, where $H=W=256$. Subsequently, we utilize 3 stacked deconvolution layers to construct a feature pyramid, with each layer $f^{\text{deconv}}_i:\mathbb{R}^{H_{\text{in}}\times W_{\text{in}}\times C_{\text{in}}}\mapsto \mathbb{R}^{2H_{\text{in}}\times 2W_{\text{in}}\times C_{\text{out}}}$ receiving input features with resolution $H_{\text{in}}=W_{\text{in}}=8,16,32$ and channel $C_{\text{in}}=512,256,256$ respectively, and producing a 256-channel feature map that is upsampled by a fator of 2. Following each deconvolution layer are Batch Normalization and ReLu activation layers. For every intermediate feature map, a specific regressor $f^{\text{reg}}_i:\mathbb{R}^{H_{\text{in}}\times W_{\text{in}}\times 256}\mapsto \mathbb{R}^{6+3}$ is tailored to embed it to $\mathbb{R}^{2000}$ and concatenates it with $\hat{\bm{R}}_o^{(i-1)},\hat{\bm{T}}_o^{(i-1)}$ to predict $\Delta\hat{\bm{R}}_o^{(i)},\Delta\hat{\bm{T}}_o^{(i)}$. Each regressor comprises two hidden Linear layers with a dimension of $1024$, as well as two output Linear layers that predict delta rotation and translation independently. Dropout layers with a probability of $0.5$ are inserted between each pair of consecutive Linear layers.

\vspace{-4mm}
\paragraph{Training.} The proposed $f^{\text{enc}},\{f^{\text{deconv}}_i\}_{i=1}^{3},\{f^{\text{reg}}_i\}_{i=1}^{3}$ are trained end-to-end with the inverse kinematics layer, supervised by $\mathcal{L} = \mathcal{L}_{\text{kp3d}} + \lambda_{\text{j2d}}\mathcal{L}_{\text{j2d}} + \mathcal{L}_{\text{twist}} + \lambda_{\text{occ-sil}}\mathcal{L}_{\text{occ-sil}} + \lambda_{\text{area}}\mathcal{L}_{\text{area}}$. Particularly, the object-oriented mesh alignment feedback loss $\mathcal{L}_{\text{maf}}=\lambda_{\text{occ-sil}}\mathcal{L}_{\text{occ-sil}} + \lambda_{\text{area}}\mathcal{L}_{\text{area}}$ is added after 55 training epochs. The loss weights are: $\lambda_{\text{j2d}}=1\times 10^{-9},\lambda_{\text{occ-sil}}=1\times 10^{-6},\lambda_{\text{area}}=2\times 10^{-7}$. The model is trained for 190 epochs on 6 NVIDIA GeForce RTX 3090 GPUs. In each epoch, we randomly sample one from 8 images to train. The training batch size is set to 8.

\vspace{-4mm}
\paragraph{Optimization.} The optimization energy function defined as $\mathcal{E}=w_{\text{visual}}\mathcal{E}_{\text{visual}} + w_{\text{imu}}\mathcal{E}_{\text{imu}}$ is configured with: $w_{\text{visual}}=20, w_{\text{imu}}=1\times 10^{5}$. We set the learning rate during optimization to $0.01$ for 30-fps data (BEHAVE~\cite{bhatnagar22behave}, InterCap~\cite{huang2022intercap} and CHAIRS~\cite{jiang2023fullbody}) and $5\times 10^{-4}$ for 60-fps data (IMHD$^2$ and HODome~\cite{zhang2023neuraldome}).

\subsection{Category-specific Motion Diffusion Filter}

\paragraph{Network Architecture.} We employ 4 transformer encoder-only layers, each equipped with 4 attention heads, to learn category-specific human-object interaction manifold. The model dimension $D_{\text{model}}=1024$ and the key, value dimension $D_{\text{key}} = D_{\text{value}} = 512$. We take $N=1000$ steps and sinusoidal positional encoding function during denoising phase. In contrast to the methodology outlined in~\cite{li2023ego}, where the condition is exactly a part of the target motion, we leverage outcomes from the preceding stage alongside raw IMU measurements as conditions to model the transition from the predictive distribution to the authentic manifold.

\vspace{-4mm}
\paragraph{Training.} We initially warm up the diffusion model solely on our complete training dataset using simple objective function for 100 epochs. Subsequently, we proceed to train the model on category-specific data, incorporating specially designed regularization terms $\mathcal{L}_{\text{consist}},\mathcal{L}_{\text{vel}}$ and $\mathcal{L}_{\text{imu}}$ to implicitly model distinct interaction patterns. The regularization term weights are $\lambda_{\text{off}} = \lambda_{\text{vel}} = \lambda_{\text{consist}} = 1,\lambda_{\text{imu}} = 100$. More detailed, we apply $\mathcal{L}_{\text{off}}$ and $\mathcal{L}_{\text{vel}}$ for 35 epochs before adding $\mathcal{L}_{\text{consist}}$ and $\mathcal{L}_{\text{imu}}$. To enhance the generation results, we maintain an exponential moving average (EMA) version of the model throughout training, updating it every 10 epochs with a decay rate of $0.995$. Additionally, we leverage Automatic Mixed Precision (AMP) to accelerate the training procedure. The model is trained for 55 epochs on a single NVIDIA GeForce RTX 3090 GPU, with the training batch size set to 128.

\begin{table*}[t!]
    \centering
    \setlength{\tabcolsep}{2.8pt}
    \begin{tabular}{cccccccc}
         \toprule
         % object & interaction & object & interaction & object & interaction & object & interaction \\
         % \cmidrule(r){1-2} \cmidrule(lr){3-4} \cmidrule(lr){5-6} \cmidrule(lr){7-8}
         & \multirow{9}{*}{\includegraphics[width=0.024\textwidth]{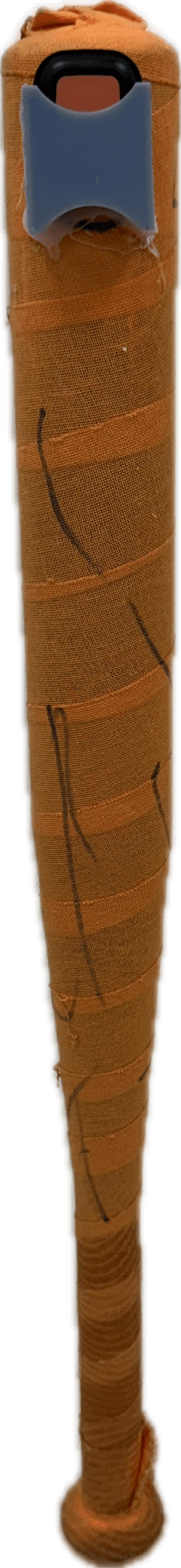}} & Holdhandle Hit & & Lefthand Carry & \multirow{9}{*}{\includegraphics[width=0.065\textwidth]{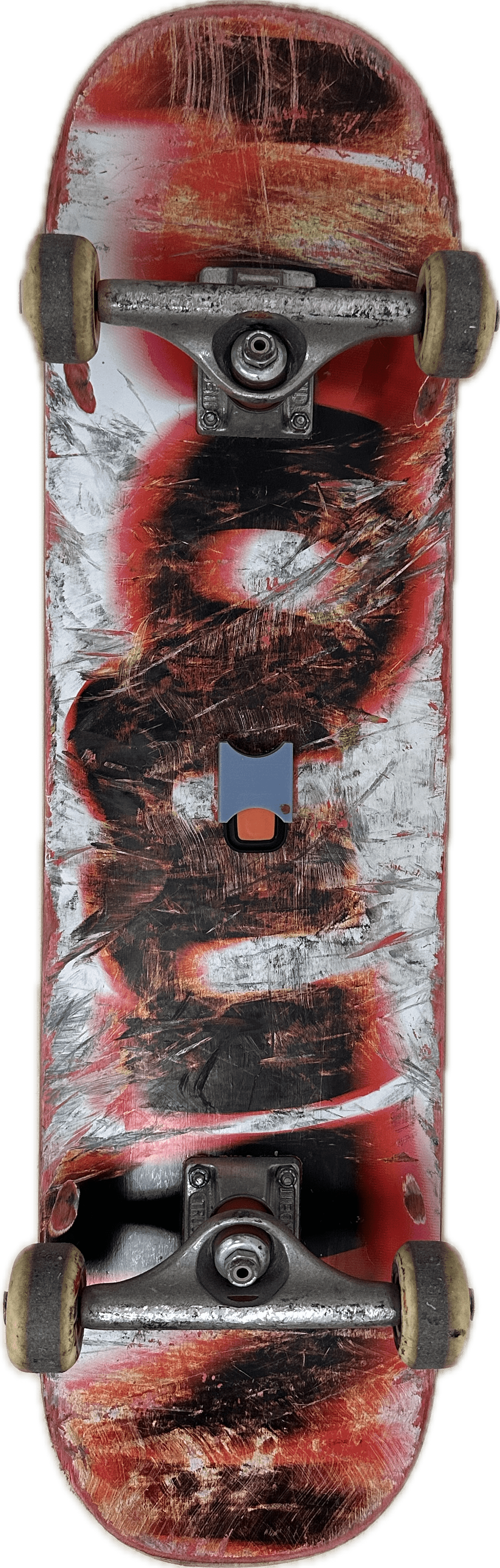}} & Ollie &  \\
         & & Holdhead Hit & \multirow{7}{*}{\includegraphics[width=0.11\textwidth]{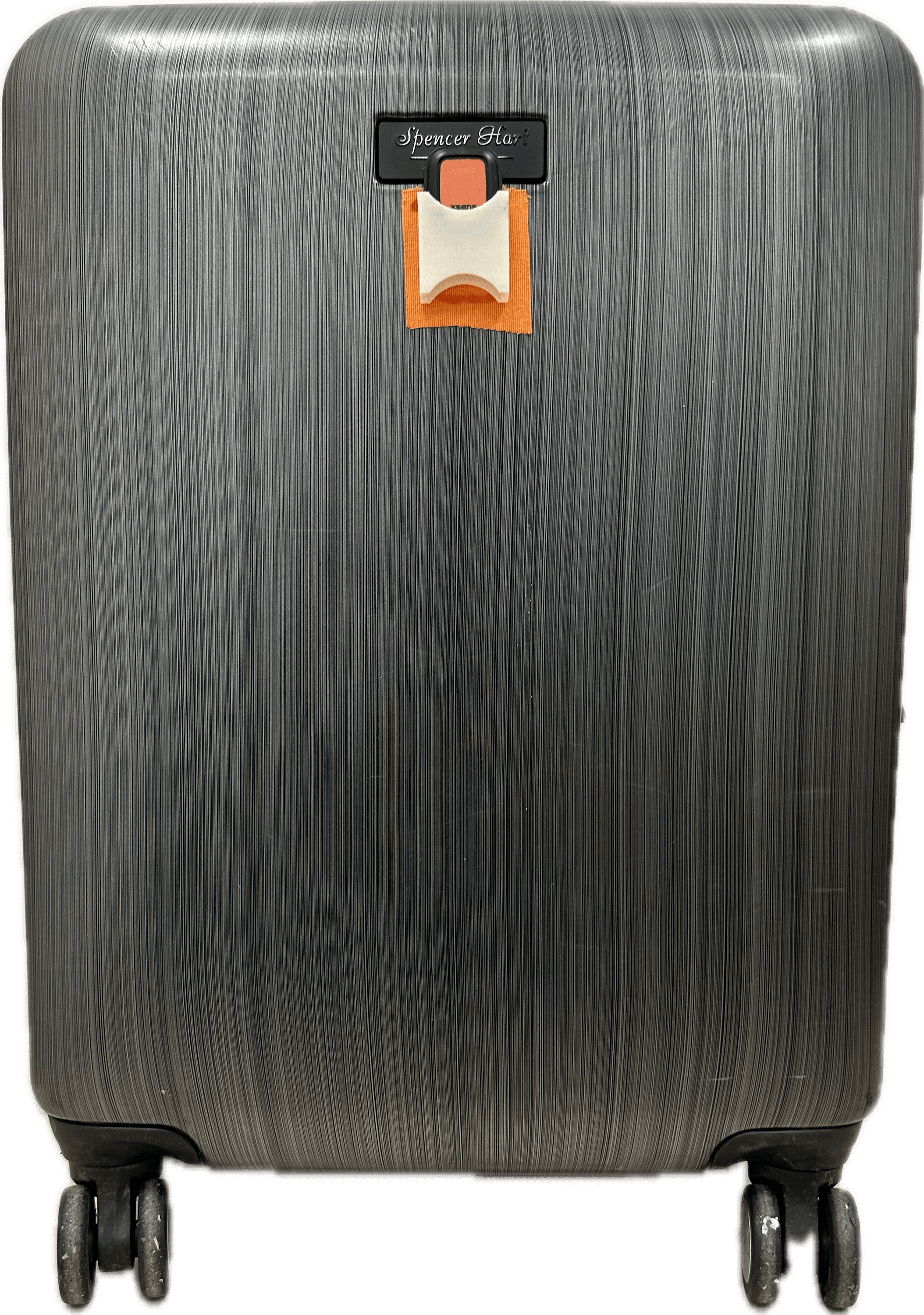}} & Lefthand Push & & Kickflip & \\
         & & Lefthand Swing & & Lift & & Grind & \\
         & & Midpart Rotate & & Putdown Pickup & & Manual & \\
         & & Pickup Putdown & & Ride Play & & Heelflip & \\
         & & Righthand Swing & & Righthand Carry & & Pop Shove-it & \\
         & & Rub  & & Righthand Push & & Nollie & \\
         & & Throw Catch  & & Twohands Carry & & Varial Kickflip & \\
         & & Twoends Rotate  & \textbf{suitcase} & Twohands Push & & McTwist & \\
         & \textbf{baseball bat} & Twohands Swing  & & Twohands Pull & \textbf{skateboard} & Darkslide & \\
         \midrule
         \multirow{5}{*}{\includegraphics[width=0.06\textwidth]{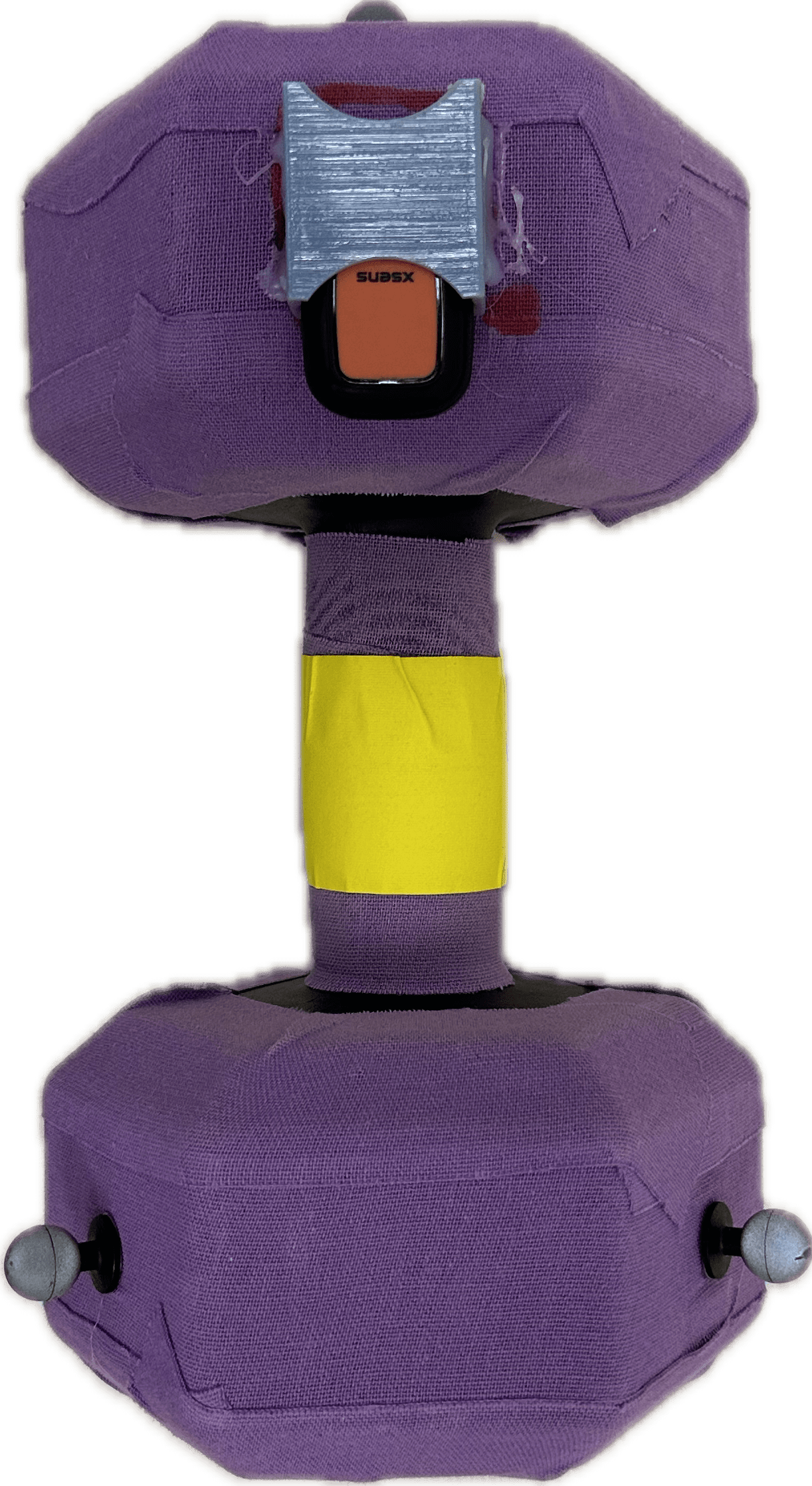}} & Left Biceps & \multirow{5}{*}{\includegraphics[width=0.08\textwidth]{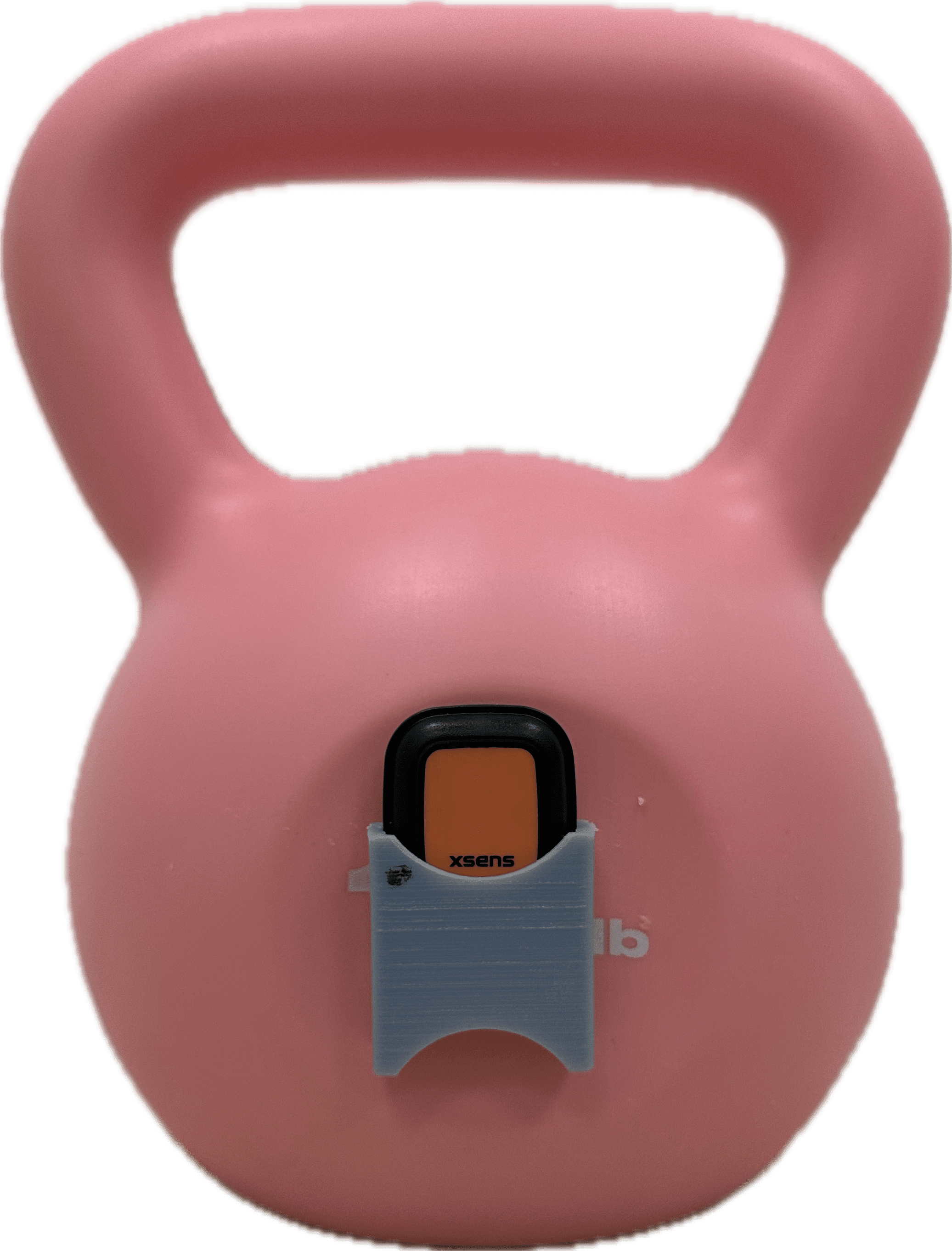}} & Forward Swing & \multirow{5}{*}{\includegraphics[width=0.045\textwidth]{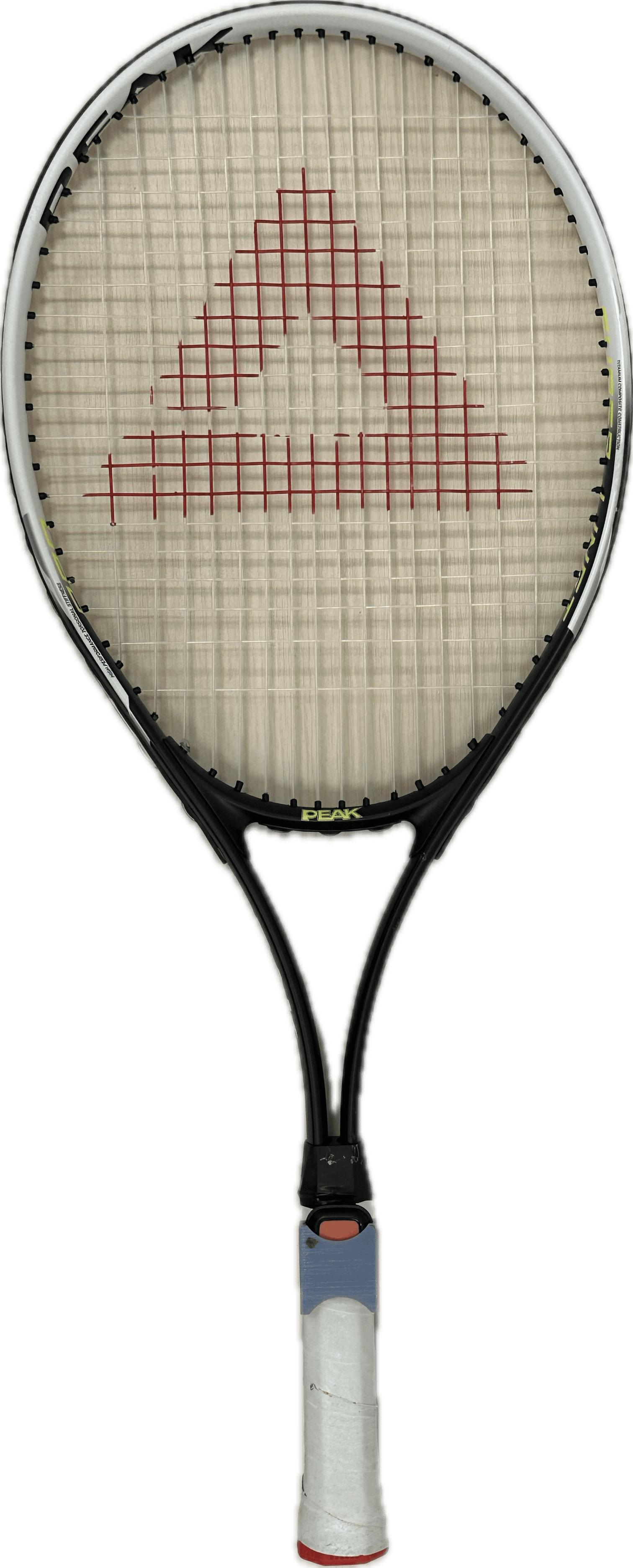}} & Forehand & \multirow{5}{*}{\includegraphics[width=0.065\textwidth]{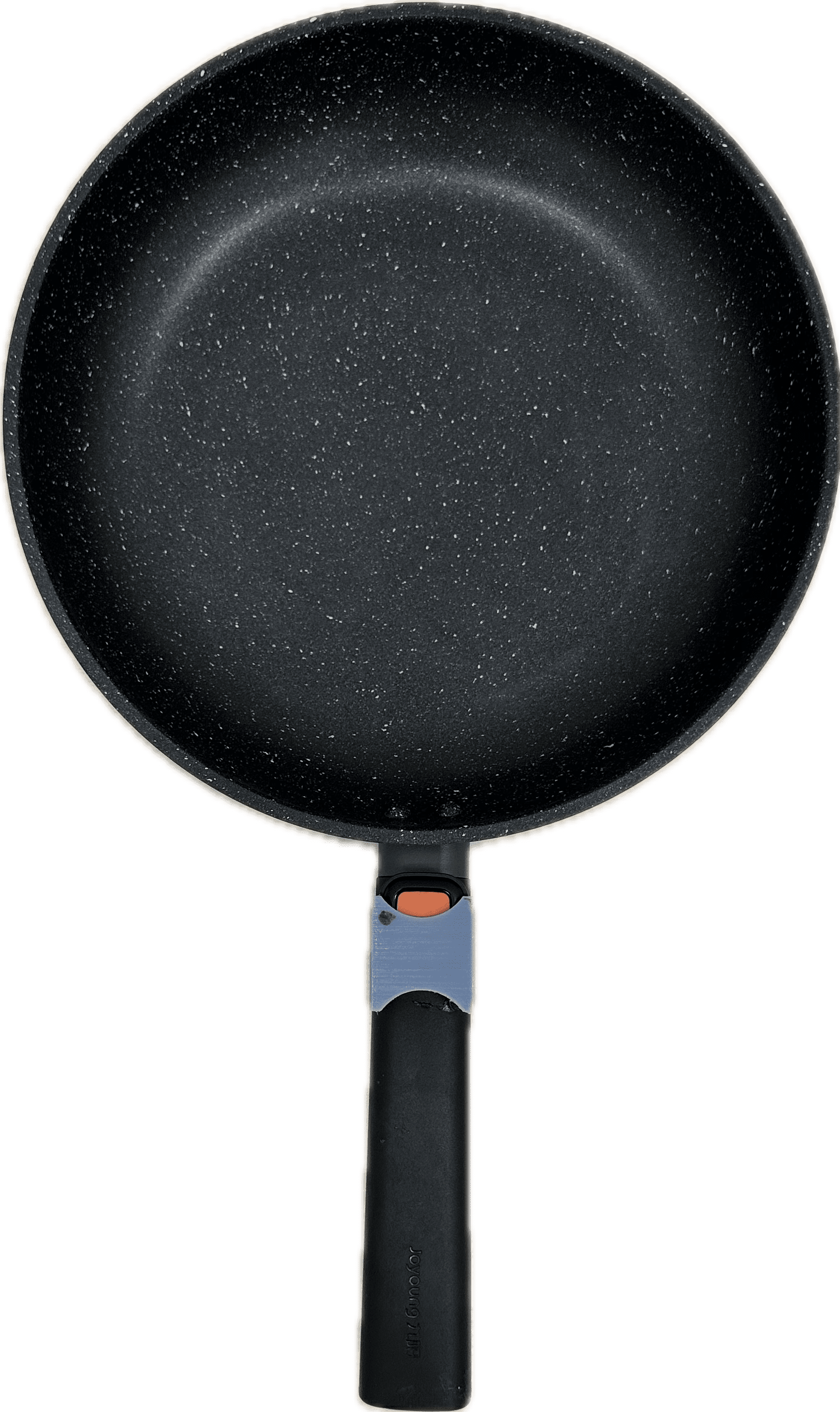}} &  \\
         & Left Lunges & & Backward Swing & & Backhand & & Hold\\
         & Left Triceps & & Snatch & & Volley & & Stir \\
         & Right Biceps & & Turkish Get-up & & Overhead Smash & & Shake \\
         & Right Lunges & & Goblet Squat & & Slice & & Flip \\
         \textbf{dumbbell} & Right Triceps & \textbf{kettlebell} & Windmill & \textbf{tennis racket} & Drop Shot & \textbf{pan} \\
         \midrule
         & \multirow{9}{*}{\includegraphics[width=0.02\textwidth]{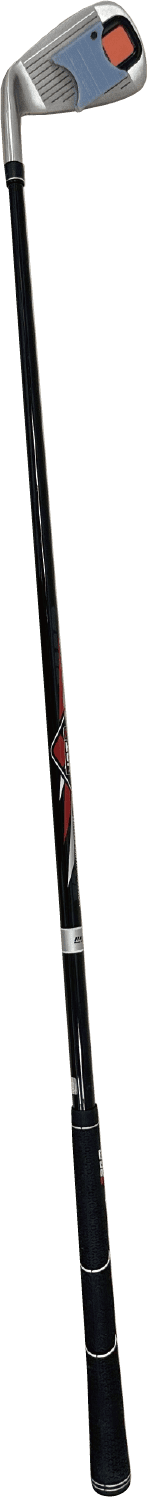}} & Drive & & Sit & \multirow{9}{*}{\includegraphics[width=0.07\textwidth]{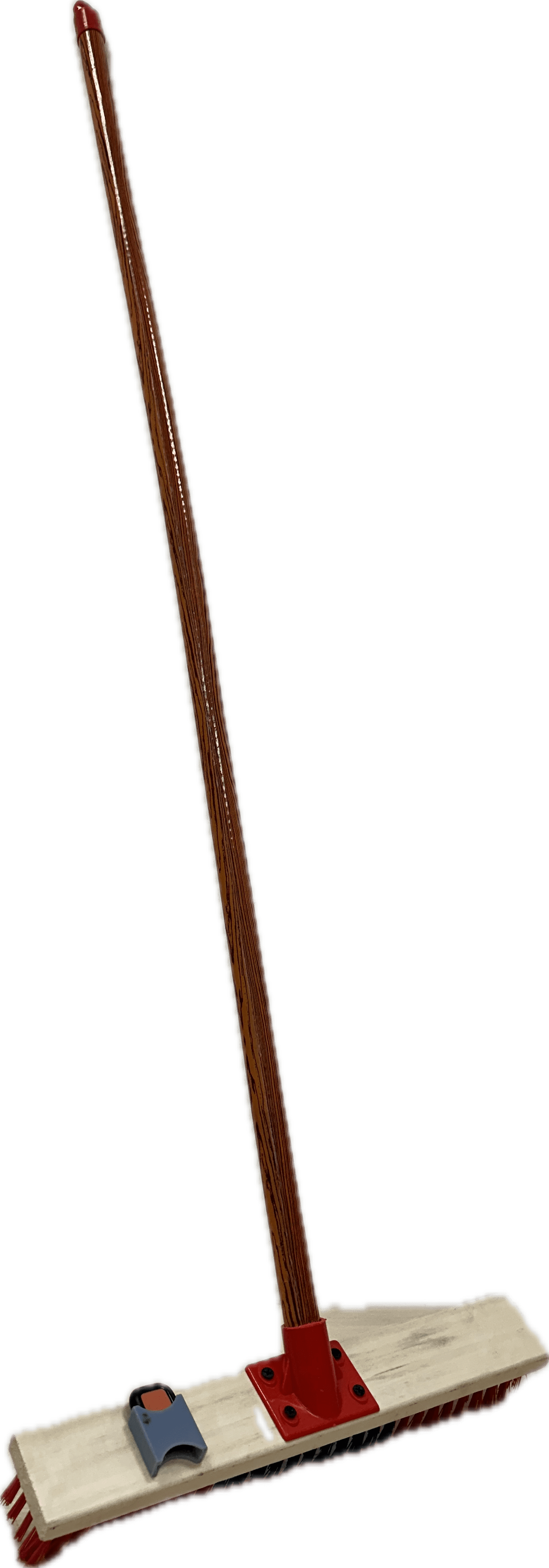}} & Sweep \\
         & & Putt & \multirow{7}{*}{\includegraphics[width=0.1\textwidth]{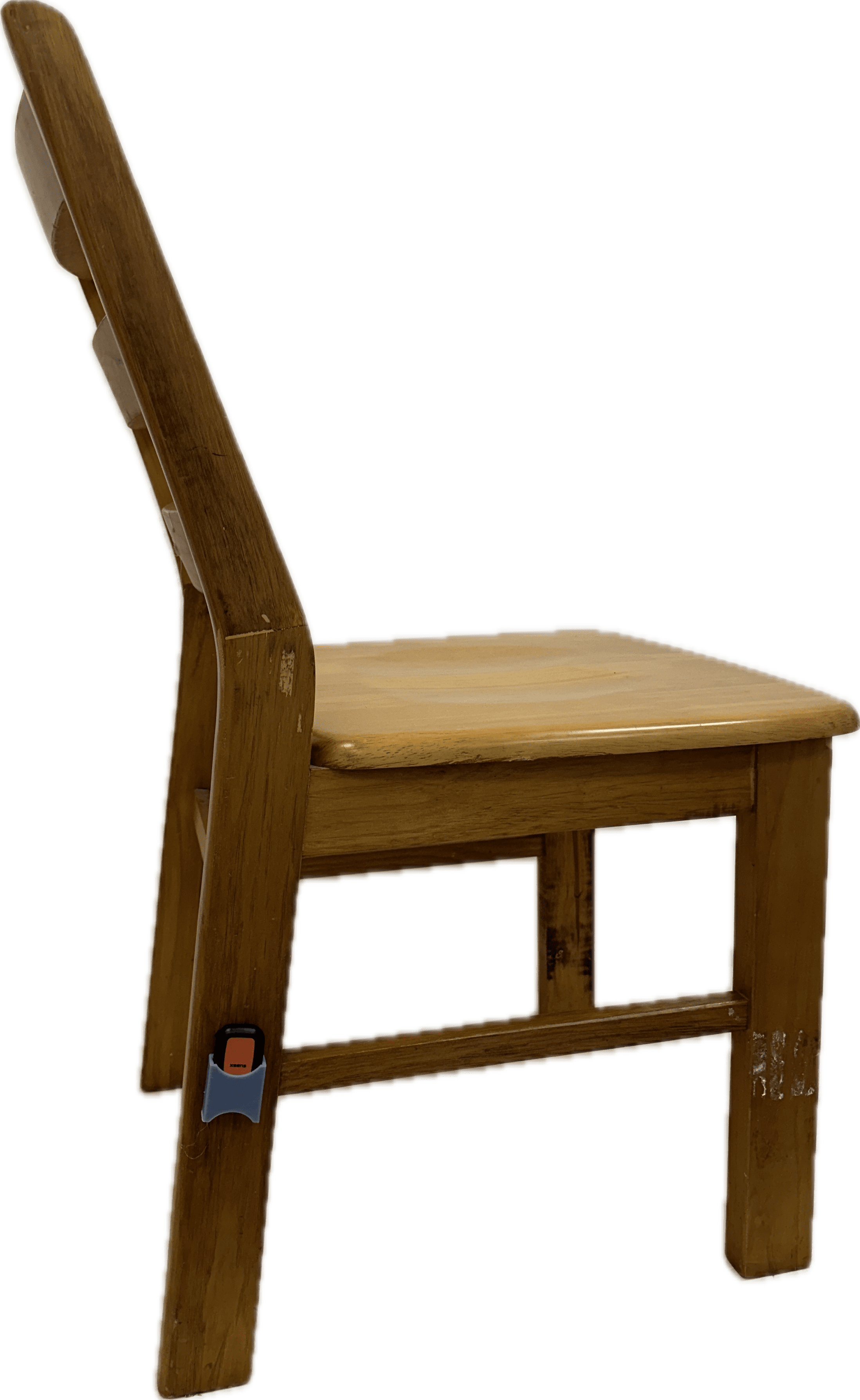}} & Lean & & Push \\
         & & Chip & & Adjust & & Pull \\
         & & Pitch & & Swivel & & Twist \\
         & & Sand Shot & & Recline & & Store \\
         & & Fade & & Rest & & Tap \\
         & & Hook & & Clean & & Tilt \\
         & & Draw & & Lift & & Lift \\
         & & Grip & \textbf{chair} & Rock & & Grip \\
         & \textbf{golf club} & Slice & & Kick & \textbf{broom} & Maintain \\
         \bottomrule
    \end{tabular}
    \caption{IMHD$^2$ collects 10 distinct objects along with a range of interaction motions associated with each object.}
    \label{tab:dataset}
\end{table*}

\section{Data Preparation Details}

\subsection{System Calibration of IMHD$^2$}

\paragraph{Temporal Synchronization.}
In order to synchronize RGB data with IMU measurements, we instructed the performer to wear an additional IMU sensor on the ankle area and execute a takeoff motion at the onset of each interaction segment. By detecting the point at which the performer falls to the ground based on the gravitational acceleration mutation in the IMU signals, we automatically pinpointed this moment as the starting frame and manually annotated it within the RGB sequences.

\vspace{-4mm}
\paragraph{Spatial Alignment.} To mitigate spatial misalignment between camera and IMU, we conducted spatial alignment once per ten minutes. Specifically, in our multi-modal and multi-sensor system, there exists multiple coordinate frames, including $\{\mathcal{F}_{C_i}\}_{i=0}^{31}$ for cameras, $\mathcal{F}_W$ for world and $\mathcal{F}_I$ for inertia. Since the transformation $\mathcal{T}_{W\rightarrow C_i}\in \mathcal{SO}(3)$ from $\mathcal{F}_W$ to $\mathcal{F}_{C_i}$ is easy to obtain through off-the-shelf multi-camera calibration toolbox, our goal is to calibrate the transformation $\mathcal{T}_{I\rightarrow W}\in \mathcal{SO}(3)$ from $\mathcal{F}_I$ to $\mathcal{F}_W$.

In our implementation, we capture the global orientation $\{\bm{R}_t^W\in \mathcal{SO}(3)\}_{t=0}^{T-1}$ of the performer who circles around in $\mathcal{F}_{W}$ by~\cite{easymocap}. The inertial rotation measurement $\{\bm{R}_t^I\in \mathcal{SO}(3)\}_{t=0}^{T-1}$ in $\mathcal{F}_{I}$ is simultaneously recorded by an IMU sensor positioned at the waist area. Suppose the IMU sensor is relatively fixed to the performer, we can construct the following equation:
\vspace{-1mm}
\begin{equation}
    \mathcal{T}_{I\rightarrow W}\bm{R}_t^I(\mathcal{T}_{I\rightarrow W}\bm{R}_{t+s}^I)^{-1} = \bm{R}_t^W(\bm{R}_{t+s}^W)^{-1},
\label{eq:calib}
\end{equation}

\vspace{-1mm}
\noindent where $s=5$ is the stride. Let $\bm{B}_t = \bm{R}_t^I(\bm{R}_{t+s}^I)^{-1}$ and $\bm{A}_t = -\bm{R}_t^W(\bm{R}_{t+s}^W)^{-1}$, we can reformulate Equation~\ref{eq:calib} as:
\vspace{-1mm}
\begin{equation}
    \bm{A}_t\mathcal{T}_{I\rightarrow W} + \mathcal{T}_{I\rightarrow W}\bm{B}_t = \bm{0},
\end{equation}

\vspace{-1mm}
\noindent which is a Sylvester equation. To solve this equation, both analytical~\cite{park1994robot} and iterative optimization methods~\cite{kingma2014adam} can be used.

\subsection{Acceleration Data Processing}

\paragraph{Normalization on Real Data.} Given the assumption that all objects are rigid and possess uniform rotational inertia, 
\begin{wrapfigure}{r}{2.8cm}
    \centering
    \vspace{-4mm}
    \includegraphics[width=0.15\textwidth]{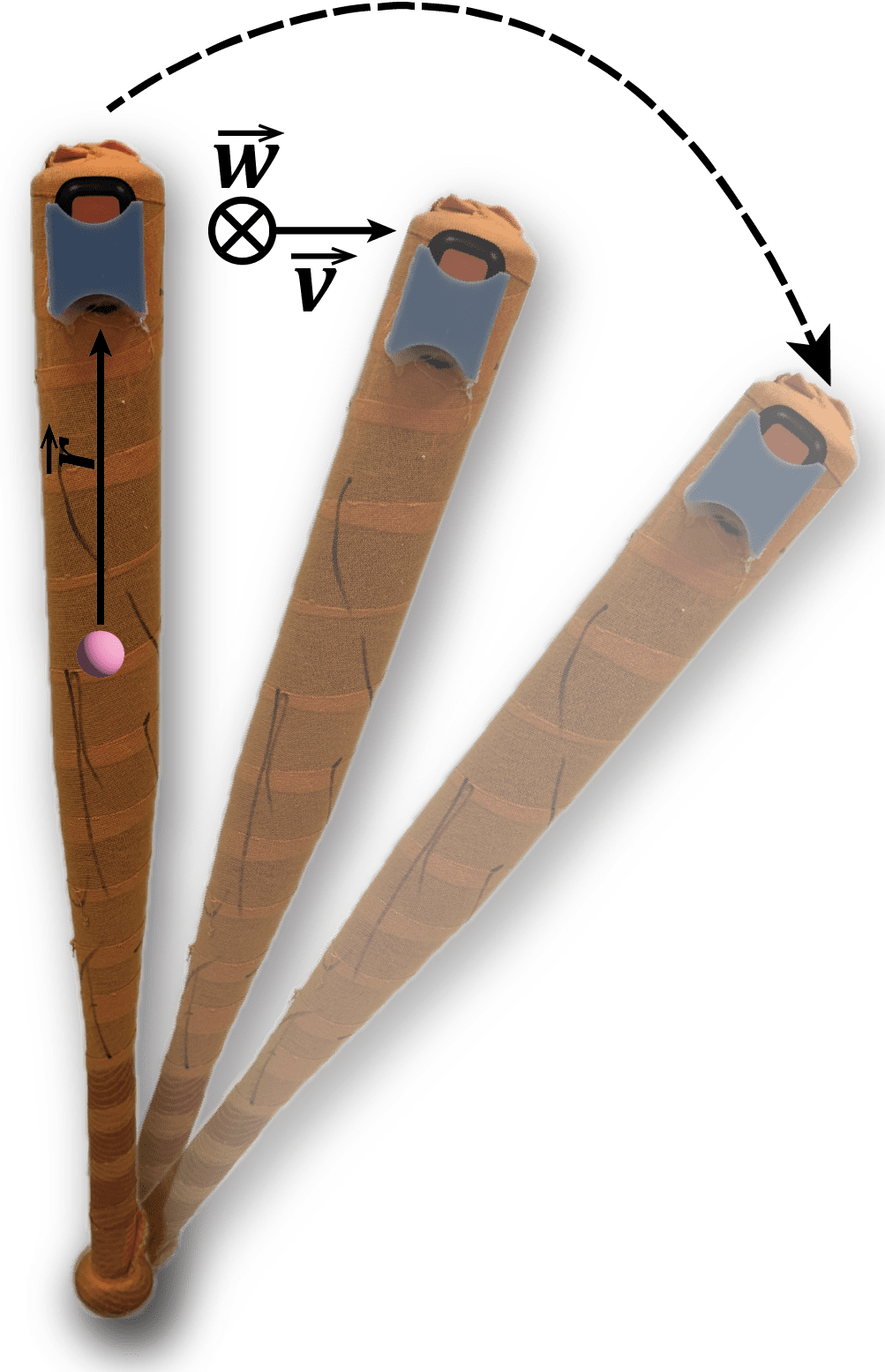}
    \vspace{-2mm}
    \caption{Illustration of why extra linear acceleration occurs.}
    \label{fig:normalization}
    \vspace{-4mm}
\end{wrapfigure}
with their centroids equivalent to their geometry centers, practical constraints arise when attempting to mount the IMU sensor precisely onto these centers. which may lie within the object. Consequently, extraneous linear acceleration may arise even from pure rotational motion, introducing undesirable noise. To eliminate such disturbances, we initially fix a mounting point for each object and manually measure the directional offset $\vec{\bm{r}}$ from the center to that point using mesh processing software~\cite{meshlab}. By leveraging recorded angular velocity $\vec{\bm{w}}$, the additional linear velocity stemming from rotation is $\vec{\bm{v}} = \vec{\bm{w}}\times \vec{\bm{r}}$, and we can calculate the excess linear acceleration: $\delta\bm{a}_t = \frac{\bm{v}_t - \bm{v}_{t-\Delta t}}{\Delta t}$. Finally, the normalized acceleration data can be attained by subtracting $\delta \bm{a}_t$ from the raw measurements.

\vspace{-4mm}
\paragraph{Simulation on Synthetic Data.} Furthermore, we simulate synthetic IMU data based on ground-truth object motion annotations of~\cite{bhatnagar22behave,huang2022intercap,zhang2023neuraldome,jiang2023fullbody}. In particular, to derive inertial acceleration data, we follow~\cite{TransPoseSIGGRAPH2021,liang2023hybridcap,ren2023lip} to calculate the second-order difference of object translation:
\begin{equation}
    \bm{a}_t = \frac{\bm{T}_{o,t-n} + \bm{T}_{o,t+n} - 2\bm{T}_{o,t}}{(n\tau)^2},
\end{equation}
where $n=4$ is the smoothing factor to enhance the approximation to actual acceleration, and $\tau=\frac{1}{\text{fps}}$ represents the time interval between consecutive frames.

\subsection{Dataset Statistics}
\begin{figure}[t!]
    \centering
    \includegraphics[width=0.45\textwidth]{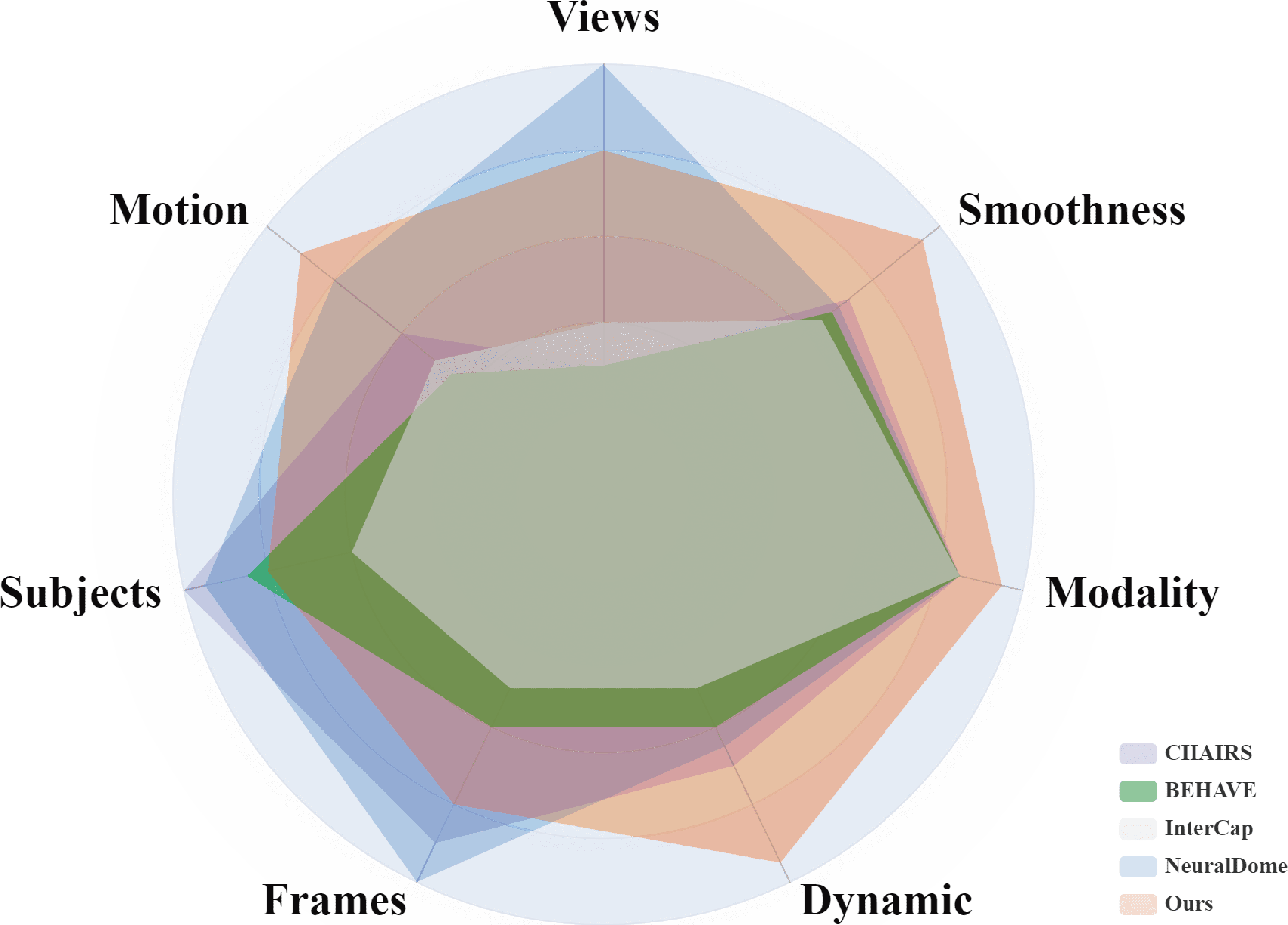}
    \caption{Attributes comparison between different datasets.}
    \label{fig:dataset_stat}
\end{figure}

We present a comprehensive overview of the contents of IMHD$^2$ in Table~\ref{tab:dataset}. It reveals that, for each object, we curated a wide array of interaction patterns involving different human body segments. Complementary to existing datasets characterized by numerous participants, extensive recording frames and super dense views, Figure~\ref{fig:dataset_stat} illustrates that IMHD$^2$ offers a more challenging, diverse and quality collection of motion data focusing on object-oriented interactions. Specifically, measured through metrics such as average motion velocity and jitter, IMHD$^2$ encompasses more dynamic interaction motions with better smoothness. Moreover, IMU data is concurrently collected alongside RGB images, serving not only to align with ground-truth annotations, but also as network input to enhance accuracy and efficiency in motion capture.

\section{More Experiments}

\subsection{More Results}
\begin{figure*}[t!]
    \centering
    \includegraphics[width=0.88\textwidth]{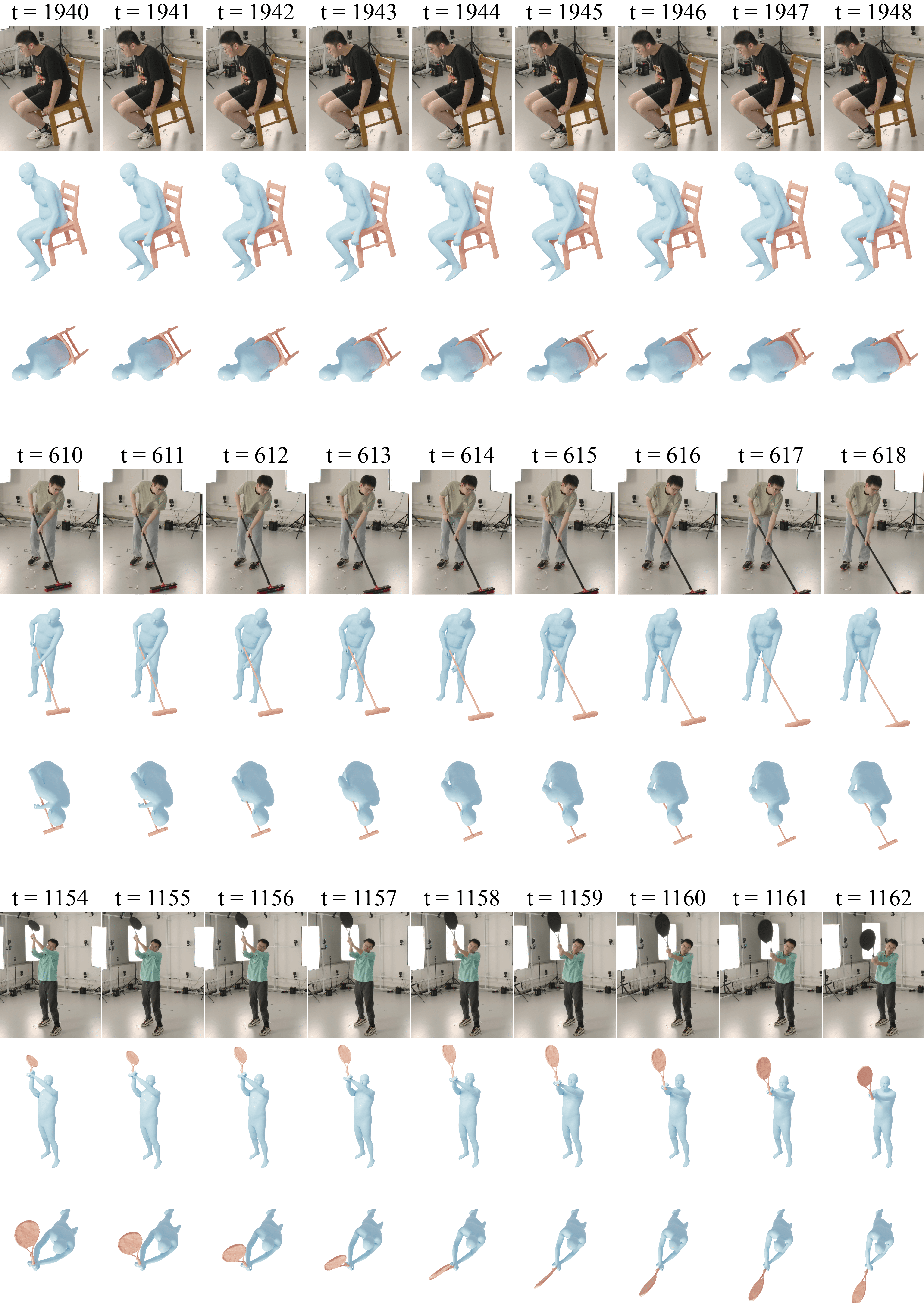}
    \caption{Additional qualitative results of I'm-HOI on IMHD$^2$. We present sequential RGB images, captured motion from camera view and top-view visualizations.}
    \label{fig:more_results}
\end{figure*}

In preceding sections, we have demonstrated the robustness of I'm-HOI under severe occlusions. Expanding on this, we now present sequential capture results of I'm-HOI to showcase spatial-temporal coherence. Figure~\ref{fig:more_results} illustrates that our approach captures accurate and consistent human-object spatial arrangements within a temporal context, which validates that our proposed network learns reasonable interaction distributions and recognizes continuous interaction behaviors from input data featuring a hybrid modality.

\subsection{More Comparisons}
\begin{figure*}[t!]
    \centering
    \includegraphics[width=0.95\textwidth]{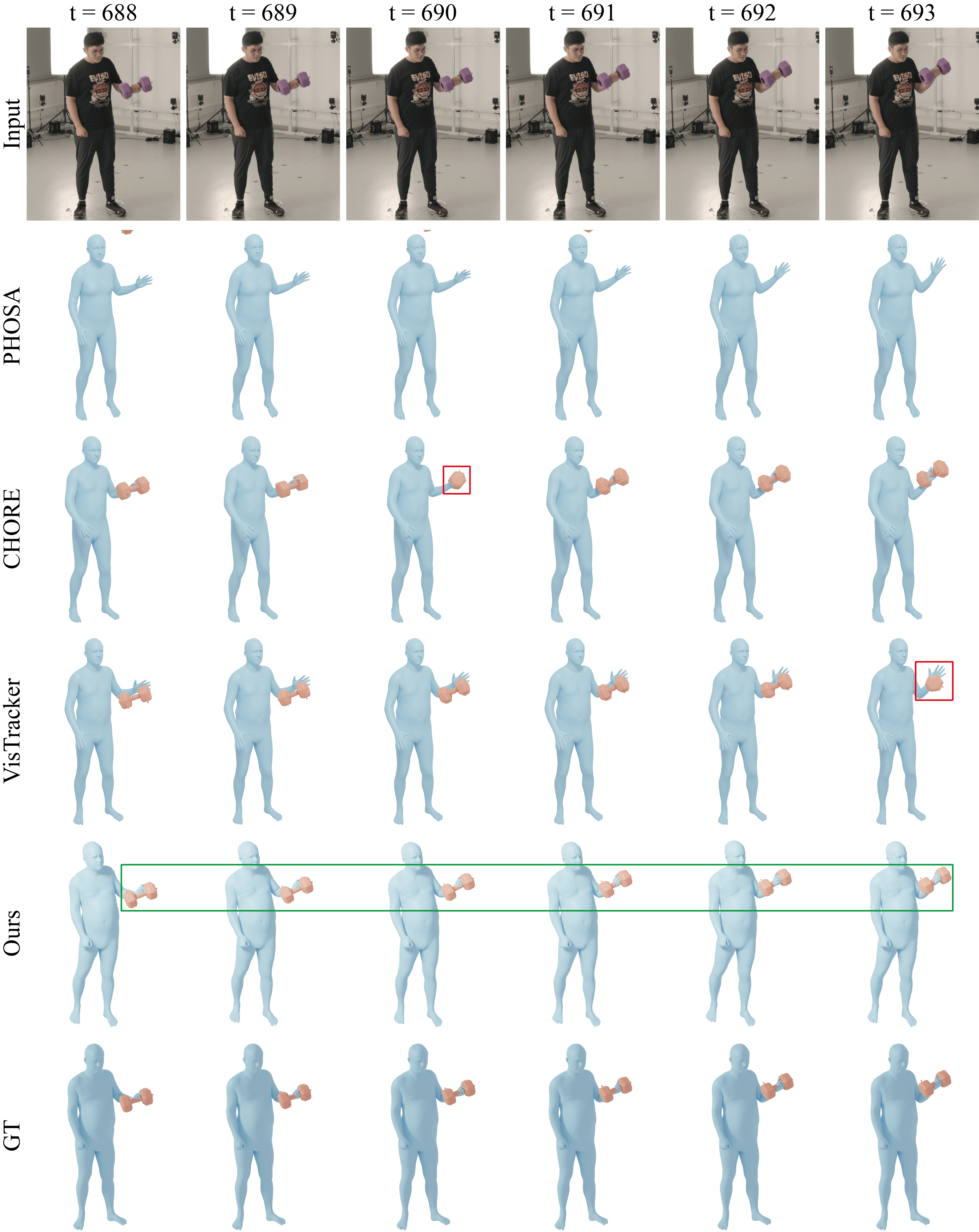}
    \caption{Additional qualitative comparisons. I’m-HOI outperforms baselines on sequential data.}
    \label{fig:more_comp1}
\end{figure*}

\begin{figure*}[t!]
    \centering
    \includegraphics[width=0.95\textwidth]{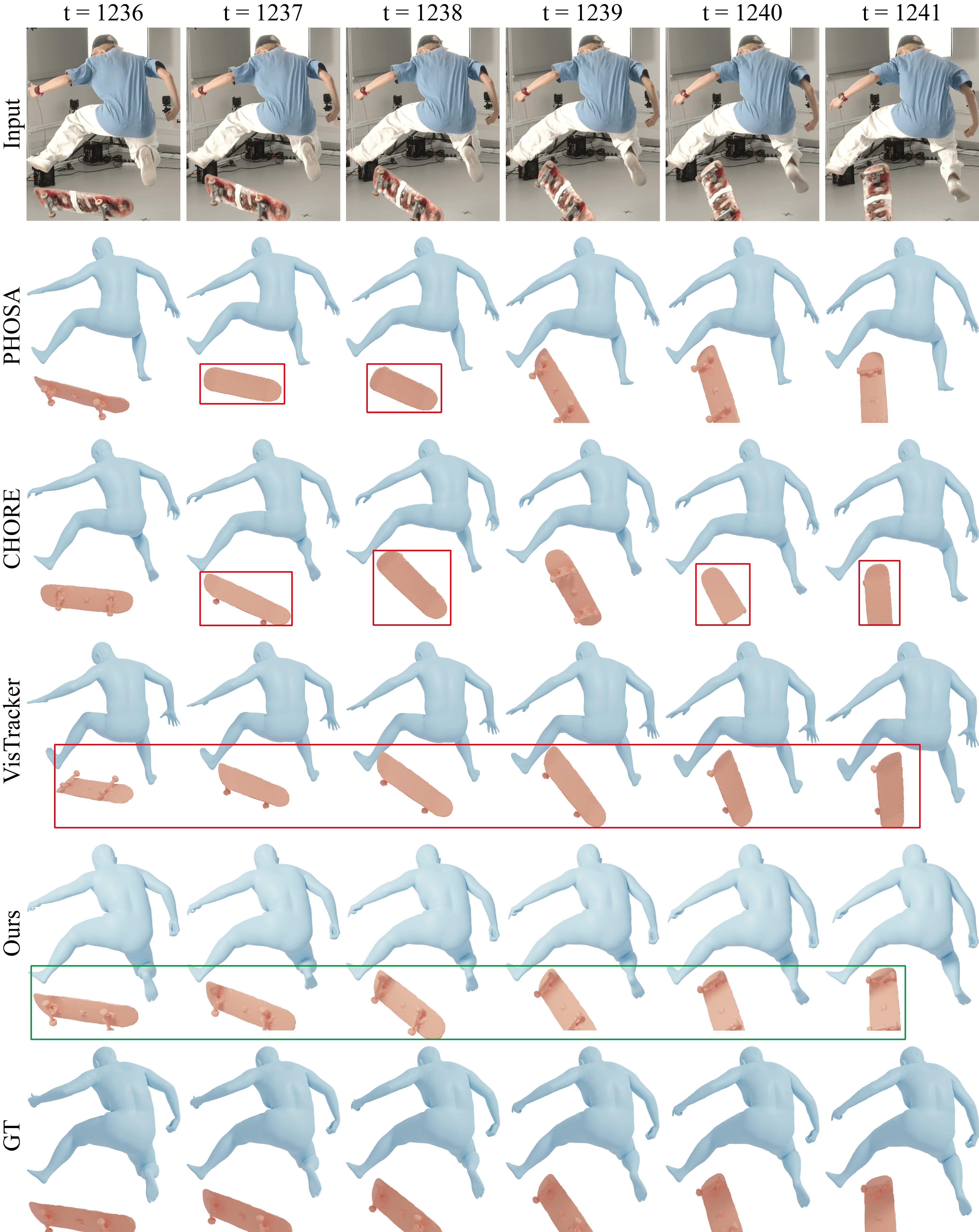}
    \caption{Additional qualitative comparisons. I’m-HOI outperforms baselines on sequential data.}
    \label{fig:more_comp2}
\end{figure*}

We also present additional qualitative comparisons of sequential capture results with baselines in Figure~\ref{fig:more_comp1} and Figure~\ref{fig:more_comp2}. It can be observed that even within an extremely short time interval (approximately 0.07 seconds), the image-based baselines~\cite{zhang2020phosa,xie2022chore} exhibit jittery object tracking results, focusing on static interactions while disregarding temporal information. Conversely, the video-based method~\cite{xie2023vistracker} yields temporally consistent but erroneous predictions without inertial measurements, particularly evident in tracking object rotational motions. In stark contrast, I'm-HOI makes use of both visual cues and IMU signals, cooperating with the design of object-oriented mesh alignment feedback and category-specific interaction prior. This combination contributes significantly to achieving consistent and correct results.

\subsection{Ablation on Regularization Terms}

\begin{table}[t!]
    \centering
    \small
    \setlength{\tabcolsep}{7pt}
    \begin{tabular}{lcccc}
        \toprule
        & \multicolumn{2}{c}{CD (per-frame)} & \multicolumn{2}{c}{CD ($10s$)}\\
        \cmidrule(lr){2-3} \cmidrule(l){4-5}
        Regularization terms & smpl & object & smpl & object \\
        \midrule
        w/o $\mathcal{L}_{\text{off}}$ & 7.07 & 7.31 & 7.59 & 9.17 \\
        w/o $\mathcal{L}_{\text{vel}}$ & 7.06 & 7.62 & 7.62 & 10.66 \\
        w/o $\mathcal{L}_{\text{consist}}$ & 6.29  & 6.98 & 6.30 & 9.01 \\
        w/o $\mathcal{L}_{\text{imu}}$ & 7.10 & 7.61 & 7.87 & 10.94 \\
        Ours & \textbf{6.50} & \textbf{6.93} & \textbf{5.36} & \textbf{8.53} \\
        \bottomrule
  \end{tabular}
  \caption{Quantitative evaluations on regularization terms.}
  \label{tab:eval_reg}
\end{table}

To further evaluate the effectiveness of the regularization terms in training of interaction diffusion filter, we conduct a comparative analysis of the full model against downgraded versions that exclude individual terms. As reported in Table~\ref{tab:eval_reg}, the inclusion of $\mathcal{L}_{\text{off}}$ restricts objects to a more specific and precise region. $\mathcal{L}_{\text{consist}}$ enforces predicted joint rotations to align with detected 3D joints after forward kinematics, which prevents overfitting to pseudo ground-truth annotations. Both $\mathcal{L}_{\text{vel}}$ and $\mathcal{L}_{\text{imu}}$ contribute to improve performance in the temporal domain. However, only applying $\mathcal{L}_{\text{vel}}$ may lead to oversmooth results due to the loss of physical dynamics. Incorporating second-order supervision $\mathcal{L}_{\text{imu}}$ is verified beneficial, not only for smooth results but also for capturing physically plausible interaction motions.

\clearpage

\end{document}